\documentclass{article}

\usepackage{microtype}
\usepackage{graphicx}
\usepackage{subfigure}
\usepackage{booktabs} 

\usepackage{amsmath}
\usepackage{etoolbox}

\usepackage{amssymb}
\usepackage{algorithmic}
\usepackage{forloop}
\usepackage{fancyhdr}
\usepackage{stmaryrd}
\usepackage{textcomp}
\usepackage{bm}

\DeclareMathOperator*\argmaxx{argmax}

\usepackage{hyperref}

\usepackage[accepted]{icml2021}

\icmltitlerunning{A New Representation of Successor Features for Transfer across Dissimilar Environments}

\begin{document}

\twocolumn[
\icmltitle{A New Representation of Successor Features for Transfer across Dissimilar Environments}



\icmlsetsymbol{equal}{*}

\begin{icmlauthorlist}
\icmlauthor{Majid Abdolshah}{Deakin}
\icmlauthor{Hung Le}{Deakin}
\icmlauthor{Thommen Karimpanal George}{Deakin}
\icmlauthor{Sunil Gupta}{Deakin}
\icmlauthor{Santu Rana}{Deakin}
\icmlauthor{Svetha Venkatesh}{Deakin}
\end{icmlauthorlist}

\icmlaffiliation{Deakin}{Applied Artificial Intelligence Institute ($\mathrm{A}^2\mathrm{I}^2$), Deakin Uni-versity, Geelong, Australia}

\icmlcorrespondingauthor{Majid Abdolshah}{majid@deakin.edu.au}

\icmlkeywords{Machine Learning, ICML}

\vskip 0.3in
]



\printAffiliationsAndNotice{ } 

\begin{abstract}
Transfer in reinforcement learning is usually achieved through generalisation
across tasks. Whilst many studies have investigated transferring knowledge
when the reward function changes, they have assumed that the dynamics
of the environments remain consistent. Many real-world RL problems
require transfer among environments with different dynamics. To address
this problem, we propose an approach based on successor features in
which we model successor feature functions with Gaussian Processes
permitting the source successor features to be treated as noisy measurements
of the target successor feature function. Our theoretical analysis
proves the convergence of this approach as well as the bounded error
on modelling successor feature functions with Gaussian Processes in
environments with both different dynamics and rewards. We demonstrate
our method on benchmark datasets and show that it outperforms current
baselines.
\end{abstract}

\section{Introduction}
Reinforcement learning (RL) is a computational approach that learns
how to attain a complex goal by maximising rewards over time. Successful
applications range from Atari games \cite{mnih2015human}, to robotics
\cite{zhang2017deep}, and self-driving cars \cite{liang2018cirl}.
However, this success is based on solving each task from scratch,
and thus training these agents requires vast amounts of data.

Several solutions have been proposed to address this problem. Most
works in transfer RL such as Progressive\textbf{ }Neural Networks\textbf{
}\cite{rusu2016progressive} and Inter-Task Mapping setups \cite{ammar2011reinforcement,gupta2017learning,konidaris2006autonomous,yin2017knowledge} 
assume that the state-action space, or reward distribution space can
be disentangled into independent sub-domains. However, learning interpretable,
disentangled representations is challenging \cite{zhu2020transfer}.
The dynamics and the reward was decoupled for the first time in \ \cite{barreto2017successor}.
Building upon an elegant formulation called the \emph{successor function}
 \cite{dayan1993improving}, the method allowed flexible transfer
learning across tasks that differ in their reward structure. The underlying
assumption was that the environmental dynamics remains unchanged and
using a Generalised Policy Improvement (GPI) method, the optimal policy
is determined. Such successor feature based methods have been shown
to efficiently transfer knowledge across RL tasks  \cite{barreto2018transfer,barreto2019option,barreto2020fast}.
Other works that have built upon successor features include generalised
policy updates on successor features \cite{barreto2020fast}, a
universal type of successor feature based on the temporal difference
method \cite{ma2020universal}, and Variational Universal Successor
Features \cite{siriwardhana2019vusfa} that perform target driven
navigation. Option Keyboard \cite{barreto2019option} leveraged
successor features to combine skills to define and manipulate options.
This allows for change in reward, but a slight change of the environment
can deteriorate the performance of a new task. If however both the
environmental dynamics and the reward differs across tasks, these
methods are unable to handle this challenge.

In real-world problems when environment dynamics and rewards change
across tasks, both these aspects need careful modelling. Failing to
do so can lead to negative transfer of knowledge from the previously
seen tasks. Thus RL methods using successor features need to be extended
to handle the changes in environmental dynamics.\textbf{} Such work
is limited. Zhang et al. \cite{zhang2017deep} aim to address this
problem by considering a linear relationship between the source and
target successor features. This modelling is restrictive and may fall
short in capturing the complexity of the changed environment dynamics.
\emph{Thus, the problem of designing a method using a successor feature
based approach to transfer knowledge from source to target environment
where the dynamics are dissimilar, is still open.}

Our new approach enables the efficient learning of novel successor
features to cater to the new target environmental dynamics. This is
done by using the distribution of the previous (source) task successor
features as a prior for the new target task. We model both the target
and source distributions through Gaussian Processes (GPs). The target
distribution is modeled as a noisy version of the source distribution.
This approach assumes that the source and target environments lie
within some proximity to each other i.e. they are similar within some
noisy envelope. However, this adjustable noisy envelope impacts the
upper bounded error on the modelling of optimal policy in the target
environment. The advantage of this approach is that the source observations
provide a head-start for the learning process and additional explorations
in the target will provide efficient convergence to the optimal policy.
We use a GPI method to estimate the target action value function.
We provide theoretical analysis and upper bounds (1) on the difference
of action-value functions when the optimal policy derived from environment
$i$ is replaced by the optimal policy derived from environment $j$;
(2) on the estimation error of the action-value function of an optimal
policy learned in a source environment when executed in the target
environment; (3) on the difference of the optimal action-value function
in the target environment and our GPI-derived action value function.
We evaluate this approach in a variety of benchmark environments with
different levels of complexity. Our key contributions are:
\begin{itemize}
\item A new method based on successor features and Gaussian Processes that
enhances transfer from source to target tasks when the dynamics of
the environment are dissimilar;
\item A theoretical analysis for the new successor based method; and,
\item An empirical comparison on diverse suit of RL benchmarks with different
levels of complexity.
\end{itemize}

\section{Background}
\subsection{Reinforcement Learning}

We model the RL framework as a Markov Decision Process (MDP) described
as $<\mathcal{S},\mathcal{A},p,R>$, where $\mathcal{S}$ is a finite
state space, $\mathcal{A}$ represents a finite action space, $p:\mathcal{S}\times\mathcal{S}\times\mathcal{A}\rightarrow[0,1]$
the transition probabilities, and $R:\mathcal{S}\times\mathcal{A}\rightarrow\mathbb{R}$
is a bounded reward function. A discount factor $\gamma\in(0,1]$
encodes the importance of future rewards with respect to the current
state. The objective of RL is to find an optimal policy $\pi(a|s):\mathcal{S}\rightarrow\mathcal{A}$
that maps the states to the actions such that it maximises the expected
discounted reward. The action-value of policy $\pi$ is defined as
$Q^{\pi}(s_{t},a_{t})=\mathbb{E}^{\pi}[\sum_{k=0}^{\infty}\ \gamma^{k}R(s_{t+k},a_{t+k})|s_{0}=s_{t},a_{0}=a_{t}]$,
where $s_{t}$ and $a_{t}$ are the state of the agent at time step
$t$, and the action that is taken in that state, respectively.

\textbf{Q-Learning:} Q-Learning is an off-policy RL approach that
aims to learn the optimal action-value function. This function can
be updated recursively as:
\begin{equation}
Q^{\pi}(s_{t},a_{t})=\mathbb{E}_{s_{t+1}}\big[R(s_{t},a_{t})+\gamma\underset{a\in\mathcal{A}}{\max}\big(Q^{\pi}(s_{t+1},a)\big)\big].
\end{equation}

After learning the action-value function, the optimal policy can be
retrieved by selecting the best action at every state: $\pi^{*}(s)\in\underset{a\in\mathcal{A}}{\argmaxx}\ Q^{*}(s,a)$.
In the next section we draw the connection between Q-Learning and
successor features.

\subsection{Successor Features \label{subsec:Successor-Features}}

The successor feature representation allows decoupling of the dynamics
of an MDP from its reward distributions. Baretto et al. \cite{barreto2017successor}
generalised the successor representations that was first formulated
in Dayan et al. \cite{dayan1993improving} decomposing the action-value
function into a set of features that encode the dynamics of the environment
and a weight that acts as a task-specific reward mapper. This decomposition
can be formulated as: 
\begin{equation}
R(s,a)={\bm{\phi}}^{\mathrm{}}(s,a)^{\mathrm{T}}{\bf w},\label{eq:reward}
\end{equation}
where ${\bm{\phi}}(s,a)\in\mathbb{R}^{D}$ are the features of $(s,a)$
that represent the dynamics of the environment and ${\bf w}$ is the
reward mapper of the environment. The two components can be learnt
through supervised learning \cite{zhu2020transfer}. Note that $\boldsymbol{\phi}(.,.)$
can be any complex model such as a neural network. Baretto et al.
\cite{barreto2017successor} showed that this decomposition can
be used in the construction of the action-value function. Let us assume
a reward function as in Eq. (\ref{eq:reward}), the action value function
can be derived as: 
\begin{align}
Q^{\pi}(s,a) & =\mathbb{E}^{\pi}\big[R(s_{t+1},a_{t+1})+\gamma R(s_{t+2},a_{t+2})+\nonumber \\
\ldots & |s_{t}=s,a_{t}=a\big]=\mathbb{E}^{\pi}\big[\sum_{t=0}^{\infty}\gamma^{t}\phi(s_{t+1},a_{t+1})\nonumber \\
 & |s_{t}=s,a_{t}=a\big]{\bf w}={\bm{\psi}}^{\mathrm{}}(s,a)^{\mathrm{T}}{\bf w},\label{eq:successor_1}
\end{align}

where ${\bm{\psi}}^{\mathrm{}}(s,a)^{\mathrm{T}}$ is\emph{ }the \emph{``Successor
Feature (SF)''} of $(s,a)$ and summarises the dynamics induced by
$\pi$. Eq. (\ref{eq:successor_1}) satisfies the Bellman Equation
and and can be learnt through any conventional method. By treating
the latent representation $\boldsymbol{\phi}(.,.)$ as the immediate
reward in the context of Q-Learning, the successor feature function
can be written as:
\begin{align}
{\bm{\psi}}^{\pi}(s,a)={\bm{\phi}}(s_{t},a_{t})+\gamma\mathbb{E}^{\pi}\big[{\bm{\psi}}^{\pi}(s_{t+1},\pi\big(s_{t+1})\big)&|s_{t}=s,\nonumber \\
&a_{t}=a\big].\label{eq:recursive_q}
\end{align}

\begin{figure}
\includegraphics[scale=0.3]{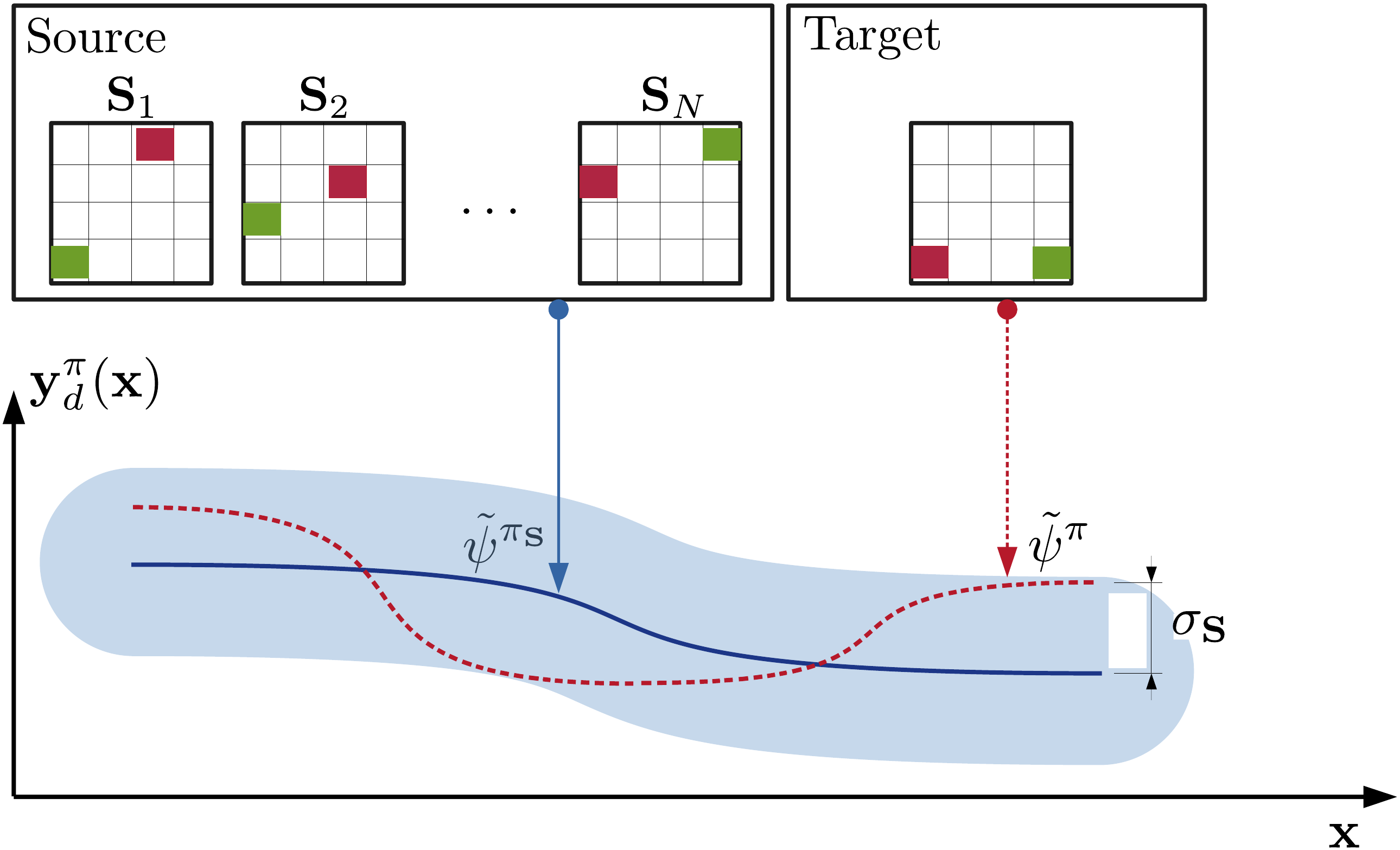}
\caption{Our proposed approach uses GPs to model the source successor features
functions ($\bm{\tilde{\psi}}^{\pi_{{\bf s}}}$) as noisy measurements
for the target successor features functions ($\bm{\tilde{\psi}}^{\pi}$).\label{fig:fig 1}}
\end{figure}
The principal advantage of SFs is that when the knowledge of ${\bm{\psi}}^{\pi}(s,a)$
is observed, one can compute a task-specific reward mapper based on
the observations seen in the same environment with different reward
function. Given ${\bm{\psi}}^{\pi}(s,a)$, the updated reward mapper
$\tilde{\boldsymbol{\mathrm{w}}}$ for a new reward function can be
approximated by solving a regression problem in tabular scenarios
\cite{barreto2020fast}. This enables us to construct the new action-value
function $\tilde{Q}^{\pi}(s,a)={\bm{\psi}}^{\pi}(s,a)^{\mathrm{T}}\boldsymbol{\tilde{\mathrm{w}}}$
by only observing few steps of the new reward function in a similar
environment. In non-tabular problems, a deep neural network can be
used to learn the successor feature functions of an environment by
minimising the following loss:
\begin{align}
L_{\psi}(\bm{\theta})=||{\bf \tilde{\bm{\phi}}}(s_{t},a_{t})+&\gamma^{t}\bm{\tilde{\psi}}^{\pi}(s_{t+1},a_{t+1};\bm{\theta}_{\psi})\nonumber\\
&-{\bf \tilde{\bm{\psi}}}^{\pi}(s_{t},a_{t};\bm{\theta}_{\psi})||.\label{eq:Loss2}
\end{align}

Similarly, by having the ${\bm{\phi}}(.,.)$ function, the new task-specific
reward mapper $\tilde{\boldsymbol{\mathrm{w}}}$ is computed as:
\begin{equation}
\mathcal{L}_{w}(\boldsymbol{\mathrm{\tilde{w}}})=\mathbb{E}_{\forall(s,a)\in\mathrm{new\ task}}\Big[\big(r(s,a)-{\bm{\phi}^{\mathrm{}}}(s,a)^{\mathrm{T}}\boldsymbol{\mathrm{\tilde{w}}}\big)^{2}\Big].\label{eq:Loss}
\end{equation}
where $r(s,a)$ is the obtained reward.

\section{Method}
We now consider source and target environments with both \emph{dissimilar
dynamics }and \emph{different reward functions}. The goal of transfer
in RL in such problems is to learn an optimal policy for the target
environment, by leveraging exterior source information and interior
target information \cite{joy2019flexible,shilton2017regret}. We
first define the set of source environments as:
\[
\mathcal{{\mathcal{M^{\mathrm{\mathbf{S}}}}}}=\Big\{\mathcal{M}(\mathcal{S},\mathcal{A},p_{1},R_{1}),\ldots,\mathcal{M}(\mathcal{S},\mathcal{A},p_{N},R_{N})\Big\},
\]

where $\mathcal{M}(.)$ represents an MDP induced by $\mathbf{\bm{\phi}}(.,.)$
as a feature function used in all environments. We denote the $N$
source environments by ${\bf S}_{1},\ldots,{\bf S}_{N}$ and train
an optimal policy in each environment to create a set of policies:
$\Pi^{\mathrm{\mathbf{S}}}=\big\{\pi^{\mathrm{\mathbf{S}}_{1}},\ldots,\pi^{\mathrm{\mathbf{S}}_{N}}\big\}$.
By executing these $N$ optimal policies in their corresponding source
environments, $N$ distinct successor feature functions ${\bf \bm{\psi}}^{\pi_{i}},\ i=\{1,\ldots,N\}$
are computed using the loss function Eq. (\ref{eq:Loss2}). We define
${\bf \bm{\psi}}^{\pi_{i}}={\bf \bm{\psi}}_{1}^{\pi_{i}},\ldots,{\bf \bm{\psi}}_{D}^{\pi_{i}}$
and ${\bf \bm{\psi}}_{d}^{\pi_{i}}$ is the $d$-th dimension. Source
successor feature functions can be learnt using neural networks with
parameters $\bm{\theta}_{\psi}$ that can be updated by~gradient
descent. Algorithm \ref{alg:Extracting-Successor-Features} shows
how the successor feature functions are computed.

Using state-action pairs as observations, we construct a set of successor
feature function samples as: $\mathcal{D}^{\mathcal{\mathbf{S}}_{i}}=\Big\{\big({\bf x}_{1}^{\mathrm{\mathbf{S}}_{i}},{\bf \bm{\tilde{\psi}}}^{\pi_{i}}({\bf x}_{1}^{\mathrm{\mathbf{\mathrm{\mathbf{S}}_{i}}}})\big),\big({\bf x}_{2}^{\mathrm{\mathrm{\mathbf{S}}_{i}}},{\bf \bm{\tilde{\psi}}}^{\pi_{i}}({\bf x}_{2}^{\mathrm{{\bf S}_{i}}})\big),\ldots,\big({\bf x}_{n}^{\mathrm{\mathrm{\mathbf{S}}_{i}}},{\bf \tilde{\bm{\psi}}}^{\pi_{i}}({\bf x}_{n}^{\mathrm{{\bf S}_{i}}})\big)\Big\},$
where ${\bf x}_{t}^{\mathrm{\mathbf{S}}_{i}}=(s_{t},a_{t})$ is a
tuple of state-action in ${\bf S}_{i}$ at time $t$ following policy
$\pi_{i}$ and $\bm{\tilde{\psi}}^{\pi_{i}}({\bf x}_{t}^{\mathrm{\mathbf{S}}_{i}})\ $
is the successor feature for ${\bf x}_{t}^{\mathrm{\mathbf{S}}_{i}}$.
If observations from target environment exist, a set of successor
feature function samples in the target environment can be constructed
as $\mathcal{D}^{\mathcal{\bm{T}}}=\Big\{\big({\bf x}_{1}^{\mathcal{\bm{T}}},{\bf \bm{\tilde{\psi}}^{\pi}}({\bf x}_{1}^{\mathcal{\bm{T}}})\big),\big({\bf x}_{2}^{\mathcal{\bm{T}}},{\bf \tilde{\bm{\psi}}}^{\pi}({\bf x}_{2}^{\mathcal{\bm{T}}})\big),\ldots,\big({\bf x}_{m}^{\mathcal{\bm{T}}},{\bf \bm{\tilde{\psi}}}^{\pi}({\bf x}_{m}^{\mathcal{\bm{T}}})\big)\Big\},$
where ${\bf x}_{t}^{\mathcal{\bm{T}}}=(s_{t},a_{t})$ records a tuple
of state-action visited at time step $t$ following policy $\pi$
in the target environment $\mathcal{T}$, and ${\bf \bm{\tilde{\psi}}^{\pi}}({\bf x}_{t}^{\mathcal{\bm{T}}})$
is an approximation of successor feature of ${\bf x}_{t}^{\mathcal{\bm{T}}}$
in the target environment. 
\begin{algorithm}[t]
\caption{Extracting Successor Feature Functions of Source Environments.\label{alg:Extracting-Successor-Features}}

\begin{algorithmic}[1]

\STATE \textbf{Input:}

\STATE Source observation and learned policies $\mathcal{D}^{\mathcal{\mathbf{S}}_{i=1...N}}=\{\},\ \Pi^{\mathrm{\mathbf{S}}}.$

\STATE Discount factor $\gamma$, exploration rate $\varepsilon_{e}$.

\STATE Feature function $\bm{\phi}(.,.)\in\mathbb{R}^{D}$.

\STATE $M_{\mathrm{max}}$ maximum number of episodes.

\STATE \textbf{Output: }Successor feature functions in source environments
$\bm{\tilde{\psi}}^{\pi_{1}},\ldots,\bm{\tilde{\psi}}^{\pi_{N}}$.

\STATE Initialise $\bm{\tilde{\psi}}^{\pi_{1...N}}:\mathcal{S}\times\mathcal{A}\rightarrow\mathbb{R}^{D}$.

\FOR {$i \in {1},\ldots,{N}$}

\WHILE {$M_\mathrm{max}$}

\STATE Sample the initial state randomly $s\in\mathcal{S}$.

\WHILE {$t\in \mathrm{steps\ not\ terminated}$}

\IF {$\varepsilon-\mathrm{greedy}$}

\STATE $a_{t}=$Uniform $(\mathcal{A}).\ \ \ \ \ $ \color{blue}
\emph{//random action}\color{black}

\ELSE

\STATE $a_{t}=\pi^{\mathrm{\mathbf{S}}_{i}}(s_{t})$.

\ENDIF

\STATE Execute $a_{t}$, observe $s_{t+1}.$

\STATE $a_{t+1}=\pi^{\mathrm{\mathbf{S}}_{i}}(s_{t+1})$.

\STATE Minimise the loss $L_{\psi^{\pi_{i}}}(\bm{\theta}).\ \ \ \ \ $\color{blue}
\emph{//Eq. (\ref{eq:Loss2})}\color{black}

\STATE Store the transition $\big[s_{t},a_{t},\bm{\tilde{\psi}}^{\pi_{i}}(s_{t},a_{t};\bm{\theta}_{\psi^{\pi_{i}}})\big]$
in $\mathcal{D}^{\mathcal{\mathbf{S}}_{i}}$.

\STATE Perform gradient descent w.r.t. $\bm{\theta}_{\psi^{\pi_{i}}}$.

\ENDWHILE

\ENDWHILE

\ENDFOR

\end{algorithmic}
\end{algorithm}

We assume that the target successor feature function ${\bf \tilde{\bm{\psi}}}_{d}^{\pi}$,
following policy $\pi$ has a measurement noise of $\epsilon{}_{d}\mathcal{\sim N}(0,\sigma^{2})$,
i.e.:
\[
{\bf y}_{d}^{\pi}({\bf x}^{\mathcal{\bm{T}}})={\bf \tilde{\bm{\psi}}}_{d}^{\pi}({\bf x}^{\mathcal{\bm{T}}})+\epsilon{}_{d},\ \forall d\in\{1,\ldots,D\},
\]
where ${\bf y}_{d}^{\pi}({\bf x}^{\mathcal{\bm{T}}})$ represents
the noisy value of successor features in the target environment. Likewise,
source successor features are assumed to have a measurement noise
of $\epsilon{}_{d}\mathcal{\sim N}(0,\sigma^{2})$, i.e.:
\small
\[
{\bf y}_{d}^{\pi_{i}}({\bf x}^{\mathrm{\mathbf{S}}_{i}})={\bf {\bf \bm{\tilde{\psi}}}}_{d}^{\pi_{i}}({\bf x}^{\mathrm{\mathbf{S}}_{i}})+\epsilon_{d},\ \forall d\in\{1,\ldots,D\},\ i=\{1,\ldots,N\}.
\]

\normalsize

We model the samples of successor features from the source environment
as noisy variants of successor features in the target environment.
Under this model, the target successor feature for observation ${\bf x}^{\mathcal{\bm{T}}}$
is defined as:
\[
{\bf \tilde{\bm{\psi}}}_{d}^{\pi}({\bf x}^{\mathcal{\bm{T}}})={\bf \tilde{\bm{\psi}}}_{d}^{\pi_{i}}({\bf x}^{\mathcal{\bm{T}}})+\epsilon{}_{d}^{\mathrm{\mathbf{S}}_{i}},\ \forall d\in\{1,\ldots,D\},
\]
where $\epsilon{}_{d}^{\mathrm{\mathbf{S}}_{i}}$ is the modelling
noise of target successor feature functions. We assume the modeling
noise to be Gaussian distributed with variance $\sigma_{\mathrm{\mathbf{S}}}^{2}$
as $\epsilon{}_{d}^{\mathrm{\mathbf{S}}_{i}}=\epsilon{}_{d}^{\mathrm{\mathbf{S}}}\mathcal{\sim N}(0,\sigma_{\mathrm{\mathbf{S}}}^{2}),\ i=\{1,\ldots,N\}$.
Intuitively, this allows the use of the source samples as a noisy
version for the target successor feature values. The value of the
``modeling noise'' variance $\sigma_{\mathrm{\mathbf{S}}}^{2}$
depends on the difference between source and target environments   \cite{shilton2017regret}. Using the successor feature samples from both source
and target environments, we model the target successor feature function
using a Gaussian Process . The overall idea is illustrated in Figure
\ref{fig:fig 1}. Without loss of generality, we use $\mathcal{GP}(0,k(.,.))$,
i.e. a GP with a zero mean function, a symmetric positive-definite
covariance function $k({\bf x},{\bf x}^{\prime}):\mathbb{X}\times\mathbb{X}\rightarrow\mathbb{R}$
a.k.a. kernel, and we also assume $k({\bf x},{\bf x})=1,\forall{\bf x}\in\mathbb{X}$.
A covariance matrix $\mathrm{\boldsymbol{\mathrm{K}}}$ based on the
combined source and target samples of successor features can be written
as:
\tiny

\begin{equation}
\mathrm{\boldsymbol{\mathrm{K}}}=\begin{bmatrix}k({\bf x}_{1}^{\mathrm{{\bf S}}},{\bf x}_{1}^{\mathrm{{\bf S}}}) & \ldots & k({\bf x}_{1}^{\mathrm{{\bf S}}},{\bf x}_{n}^{\mathrm{{\bf S}}}) & k({\bf x}_{1}^{\mathrm{{\bf S}}},{\bf x}_{1}^{\mathcal{\bm{T}}}) & \ldots & k({\bf x}_{1}^{\mathrm{{\bf S}}},{\bf x}_{m}^{\mathcal{\bm{T}}})\\
\vdots & \ddots & \vdots & \vdots & \vdots & \vdots\\
k({\bf x}_{n}^{\mathrm{{\bf S}}},{\bf x}_{1}^{\mathrm{{\bf S}}}) & \ldots & k({\bf x}_{n}^{\mathrm{{\bf S}}},{\bf x}_{n}^{\mathrm{{\bf S}}}) & k({\bf x}_{n}^{\mathrm{{\bf S}}},{\bf x}_{1}^{\mathcal{\bm{T}}}) & \vdots & k({\bf x}_{n}^{\mathrm{{\bf S}}},{\bf x}_{m}^{\mathcal{\bm{T}}})\\
k({\bf x}_{1}^{\mathcal{\bm{T}}},{\bf x}_{1}^{{\bf S}}) & \ldots & k({\bf x}_{1}^{\mathcal{\bm{T}}},{\bf x}_{n}^{{\bf S}}) & k({\bf x}_{1}^{\mathcal{\bm{T}}},{\bf x}_{1}^{\mathcal{\bm{T}}}) & \vdots & k({\bf x}_{1}^{\mathcal{\bm{T}}},{\bf x}_{m}^{\mathcal{\bm{T}}})\\
\vdots & \vdots & \vdots & \vdots & \ddots & \vdots\\
k({\bf x}_{m}^{\mathcal{\bm{T}}},{\bf x}_{1}^{{\bf S}}) & \ldots & k({\bf x}_{m}^{\mathcal{\bm{T}}},{\bf x}_{n}^{{\bf S}}) & k({\bf x}_{m}^{\mathcal{\bm{T}}},{\bf x}_{1}^{\mathcal{\bm{T}}}) & \ldots & k({\bf x}_{m}^{\mathcal{\bm{T}}},{\bf x}_{m}^{\mathcal{\bm{T}}})
\end{bmatrix},\normalsize\label{eq:Kernel}
\end{equation}

\normalsize

where $k({\bf x}_{i}^{\mathrm{{\bf S}}},{\bf x}_{j}^{\mathrm{{\bf S}}}),\ i,j=\{1,...,n\}$
is the self-covariance among source observations of ${\bf S}\in\{{\bf S}_{1},\ldots,{\bf S}_{N}\}$
and $k({\bf x}_{i}^{\mathrm{{\bf S}}},{\bf x}_{j}^{\mathcal{\bm{T}}}),\ i=\{1,...,n\},j=\{1,...,m\}$
denotes the covariance between source and target observations. After
incorporating the source, target measurement noise with the modeling
noise, the covariance matrix can be written as:
\begin{equation}
\mathrm{\mathrm{\boldsymbol{\mathrm{K}}}_{*}}=\boldsymbol{\mathrm{K}}+\begin{bmatrix}(\sigma_{{\bf S}}^{2}+\sigma_{\mathcal{}}^{2})\boldsymbol{\mathrm{I}}_{n\times n} & \boldsymbol{0}\\
\boldsymbol{0} & \sigma_{\mathcal{}}^{2}\mathrm{\boldsymbol{I}}_{m\times m}
\end{bmatrix}.\label{eq:KStar}
\end{equation}
Intuitively, a higher value of $\sigma_{{\bf S}}^{2}$ implies higher
uncertainty about similarity between source and target environments.
Having defined the required components, using the property of GP \cite{rasmussen2003gaussian},
the predictive mean and variance for a new target observation ${\bf x}^{\mathcal{\bm{T}}}=(s,a)$
is derived as:
\begin{align}
\mu_{m,d}({\bf x}^{\mathcal{\bm{T}}}) & ={\bf k}^{\mathrm{T}}\mathrm{\boldsymbol{\mathrm{K}}}_{*}^{-1}{\bf y}_{d},\label{eq:MeanP}\\
\sigma_{m,d}^{2}({\bf x}^{\mathcal{\bm{T}}}) & =k({\bf x}^{\mathcal{\bm{T}}},{\bf x}^{\mathcal{\bm{T}}})-{\bf k}^{\mathrm{T}}\mathrm{\boldsymbol{\mathrm{K}}}_{*}^{-1}{\bf k},\label{eq:GP}
\end{align}
where $\boldsymbol{\mathrm{K}}_{*}$ is the kernel matrix as defined
in Eq. (\ref{eq:KStar}) and ${\bf k}=[k({\bf x}_{i},{\bf x}^{\mathcal{\bm{T}}})],\forall{\bf x}_{i}\in\mathcal{D}^{\mathcal{\mathbf{S}}}\bigcup\mathcal{D^{\bm{T}}}$.
We use the posterior mean as in Eq. (\ref{eq:MeanP}) as the predicted
value of the $d$-th successor feature dimension for the target observation
as ${\bf \tilde{\bm{\psi}}}_{d}^{\pi}({\bf x}^{\mathcal{\bm{T}}}))=\mu_{m,d}({\bf x}^{\mathcal{\bm{T}}}).$

Using GPI \cite{barreto2017successor}, we identify the optimal
policy by selecting the best action of the best policy as:
\begin{equation}
\pi^{\prime}(s)=\underset{a\in\mathcal{A}}{\argmaxx\ }\underset{\pi\mathcal{\in}\Pi^{\mathrm{\mathbf{S}}}}{\mathrm{max}}\tilde{Q}^{\pi}(s,a).\label{eq:GPI}
\end{equation}
We note that $\tilde{Q}^{\pi}({\bf x}^{\mathcal{\bm{T}}})\approx{\bf \tilde{\bm{\psi}}}^{\pi}({\bf x}^{\mathcal{\bm{T}}})^{\mathrm{T}}\tilde{\boldsymbol{w}}$,
where $\tilde{\boldsymbol{w}}$ is obtained by minimising the loss
function in Eq. (\ref{eq:Loss}) as the agent interacts with the target
environment. We term our method \emph{Successor Features for Dissimilar
Environments (SFDE) }and\emph{ }is detailed in Algorithm \ref{alg:Successor-Features-for}.

\subsection{Theoretical Analysis}

This section answers the key question of ``what are the effects of
relaxing the assumption of similarity among source and target environments
on the convergence and transfer via successor features?''. We prove
(1) an upper bound on the difference of action-value functions when
the optimal policy derived from environment $i$ is replaced by the
optimal policy derived from environment $j$ (\emph{Theorem 1});
(2) an upper bound on the estimation error  of the action-value function
of an optimal policy learned in $\boldsymbol{\mathrm{S}}_{j}$ when
executed in target environment $\mathcal{T}$ (\emph{Lemma 1}); (3)
Using (1) and (2) an upper bound on the difference of the optimal
action-value function in the target environment and our GPI-derived
action value function (\emph{Theorem 2}).

\subsection*{Theorem 1}

Let ${\bf S}_{i}$ and ${\bf S}_{j}$ be two different source environments
with dissimilar transition dynamics $p_{i}$ and $p_{j}$ respectively.
Let $\delta_{ij}\triangleq\mathrm{max}_{s,a}\ |r_{i}(s,a)-r_{j}(s,a)|$,
where $r_{i}(.,.)$ and $r_{j}(.,.)$ are the reward functions of
environment ${\bf S}_{i}$ and ${\bf S}_{j}$ respectively. We denote
$\pi_{i}^{*}$ and $\pi_{j}^{*}$ as optimal policies in ${\bf S}_{i}$
and ${\bf S}_{j}$. It can be shown that the difference of their action-value
functions is upper bounded as:\small
\begin{align}
{Q}_{i}^{\pi_{i}^{*}}(s,a)-{Q}_{i}^{\pi_{j}^{*}}(s,a) & \leq\frac{2\delta_{ij}}{1-\gamma}\nonumber \\
+ & \frac{\gamma\Big|\Big|{\bf P}_{i}(s,a)-{\bf P}_{j}(s,a)\Big|\Big|}{(1-\gamma)}\nonumber \\
\times & \frac{\Big(\Big|\Big|{\bf Q}_{i}^{i}-{\bf Q}_{j}^{j}\Big|\Big|+\Big|\Big|{\bf Q}_{j}^{j}-{\bf Q}_{i}^{j}\Big|\Big|\Big)}{(1-\gamma)},\label{eq:eq:Inequality}
\end{align}
\normalsize

where ${Q}_{i}^{\pi_{k{\bf }}^{*}}$ shows the action-value function
in environment ${\bf S}_{i}$ by following an optimal policy that
is learned in the environment ${\bf S}_{k}\in\{{\bf S}_{1},\ldots,{\bf S}_{N}\}$.
We also define ${\bf P}_{i}(s,a)=[p_{i}(s^{\prime}|s,a),...]_{\forall s^{\prime}\in\mathcal{S}}$,
${\bf P}_{j}(s,a)=[p_{j}(s^{\prime}|s,a),...]_{\forall s^{\prime}\in\mathcal{S}},$
${\bf Q}_{i}^{i}=[\underset{b\in\mathcal{A}}{\mathrm{max}}{Q}_{i}^{\pi_{i}^{*}}(s^{\prime},b),...]_{\forall s^{\prime}\in\mathcal{S}},$
${\bf Q}_{j}^{j}=[\underset{b\in\mathcal{A}}{\mathrm{max}}{Q}_{j}^{\pi_{j}^{*}}(s^{\prime},b),...]_{\forall s^{\prime}\in\mathcal{S}}$,
${\bf Q}_{i}^{j}=[\underset{b\in\mathcal{A}}{\mathrm{max}}{Q}_{i}^{\pi_{j}^{*}}(s^{\prime},b),...]_{\forall s^{\prime}\in\mathcal{S}}$,
$\gamma$ as the discount factor, and $||.||$ to be $2-$norm (Euclidean
norm).

\textbf{Proof: }We provide a sketch of the proof which involves two
key steps. The left side of the inequality (\ref{eq:eq:Inequality})
can be rewritten as:
\small
\begin{align*}
{Q}_{i}^{\pi_{i}^{*}}(s,a)-{Q}_{i}^{\pi_{j}^{*}}(s,a) & ={Q}_{i}^{\pi_{i}^{*}}(s,a)-{Q}_{j}^{\pi_{j}^{*}}(s,a)\\
+ & {Q}_{j}^{\pi_{j}^{*}}(s,a)-{Q}_{i}^{\pi_{j}^{*}}(s,a)\\
\leq & \underbrace{|{Q}_{i}^{\pi_{i}^{*}}(s,a)-{Q}_{j}^{\pi_{j}^{*}}(s,a)|}_{\text{(I)}}\\
+ & \underbrace{|{Q}_{j}^{\pi_{j}^{*}}(s,a)-{Q}_{i}^{\pi_{j}^{*}}(s,a)|}_{\text{(II)}}.
\end{align*}

\normalsize

We can prove that $\mathrm{(I)}\leq\frac{\delta_{ij}}{1-\gamma}+\gamma\Big|\Big|{\bf P}_{i}-{\bf P}_{j}\Big|\Big|\times\Big|\Big|{\bf Q}_{i}^{i}-{\bf Q}_{j}^{j}\Big|\Big|/(1-\gamma)$
and $\mathrm{(II)}\leq\frac{\delta_{ij}}{1-\gamma}+\gamma\Big|\Big|{\bf P}_{i}-{\bf P}_{j}\Big|\Big|\times\Big|\Big|{\bf Q}_{j}^{j}-{\bf Q}_{i}^{j}\Big|\Big|\Big)/(1-\gamma$),
leading to the upper bound. Detailed proof is available in supplementary
material.

\begin{algorithm}[t]
\caption{Successor Features for Dissimilar Environments.\label{alg:Successor-Features-for}}

\begin{algorithmic}[1]

\STATE \textbf{Input:}

\STATE Source environments $\mathcal{D}^{\mathcal{\mathbf{S}}_{1...N}}$
and target observations $\mathcal{D^{\bm{T}}}=\{\}.$

\STATE Set the amount of noises for source and target $\sigma_{{\bf S}}^{2},\sigma^{2}.$

\STATE SFs of source environments $\bm{\tilde{\psi}}^{\pi_{1}},\ldots,\bm{\tilde{\psi}}^{\pi_{N}}$.

\STATE Feature function $\bm{\phi}(.,.)\in\mathbb{R}^{D}$.

\STATE Initialise reward mapper weight for target environment $\boldsymbol{\tilde{w}}$.

\STATE \textbf{Output: }Optimal policy $\pi^{*}(s)$ for the target
environment.

\WHILE {$t\in \mathrm{steps\ not\ terminated}$}

\FOR {$\forall \mathrm{\bf S}\in \{\mathrm{\bf S}_1,\ldots,\mathrm{\bf S}_N\}$}

\FOR {$a^\prime \in \mathcal{A}$}

\STATE ${\bf x}_{t}^{\mathcal{\bm{T}}}=(s_{t},a^{\prime}).$

\STATE $\mathcal{GP}_{d}^{\mathcal{\mathbf{S}}}(\{\mathcal{D}^{\mathcal{\mathbf{S}}},\mathcal{D}^{\bm{\mathcal{T}}}\}),\ \forall d=\{1,\ldots,D\}.\ $
\color{blue}{\small //Fit the GPs with source and target data} \color{black}

\STATE Calculate ${\mathrm{\boldsymbol{\mathrm{K}}}_{*}},\mathrm{\boldsymbol{k}.\ }$
\color{blue}//(Eq. 8) \color{black}

\STATE ${\bf \tilde{\bm{\psi}}}_{d}^{\pi_{\mathrm{{\bf S}}}}({\bf {\bf x}}_{t}^{\mathcal{T}}))=\mu_{m,d}^{\mathrm{{\bf S}}}({\bf x}_{t}^{\mathcal{\bm{T}}}).\ \forall d=\{1,\ldots,D\}.\ $
\color{blue}//(Eq. 9) \color{black}

\ENDFOR

\ENDFOR

\STATE Reconstruct the action-value functions of all source environments
given the target observations $\tilde{Q}^{\pi_{i}}(s_{t},a^{\prime})=\bm{\tilde{\psi}}^{\pi_{i}}(s_{t},a^{\prime})\boldsymbol{\tilde{w}},\forall a^{\prime}\in\mathcal{A},\forall i=\{1,\ldots,N\}.$

\STATE $\pi^{*}(s)=\underset{a^{\prime}\in\mathcal{A}}{\argmaxx\ }\underset{\pi\mathcal{\in}\Pi^{\mathrm{\mathbf{S}}}}{\mathrm{max}}\tilde{Q}^{\pi}({\bf x}_{t}^{\mathcal{\bm{T}}}).\ $
\color{blue}//Performing GPI \color{black}

\STATE Add the new target observation to $\mathcal{D}^{\bm{\mathcal{T}}}.$

\STATE Update the estimation of $\boldsymbol{\tilde{w}}$ based on
Eq. (\ref{eq:Loss}).

\ENDWHILE

\end{algorithmic}
\end{algorithm}
Theorem 1  uses $\delta_{ij}$ as a metric of maximum immediate reward
dissimilarities in the environments ${\bf S}_{i}$ and ${\bf S}_{j}$.
Clearly, the higher $\delta_{ij}$, the less similar ${\bf S}_{i}$
and ${\bf S}_{j}$ are, and the upper bound will be looser accordingly.
This upper bound also depends on the value of $\Big|\Big|{\bf P}_{i}-{\bf P}_{j}\Big|\Big|$
that captures the difference in dynamics of ${\bf S}_{i}$ and ${\bf S}_{j}$
- larger the value, looser the upper bound. However, $\Big|\Big|{\bf Q}_{i}^{i}-{\bf Q}_{j}^{j}\Big|\Big|+\Big|\Big|{\bf Q}_{j}^{j}-{\bf Q}_{i}^{j}\Big|\Big|$
incorporates the difference of action-value functions that are related
to both the dynamics and the future discounted reward in the two environments.
Hence, a larger value of this term implies that ${\bf S}_{i}$ and
${\bf S}_{j}$ are expected to produce different sum of discounted
future reward by following their corresponding policies. Note that
if ${\bf S}_{i}={\bf S}_{j}$ (in terms of both dynamics and reward),
the upper bound will vanish to $0$ as the two environments are identical.
Clearly, our bound is an extension of the bound in \cite{barreto2018transfer}
for environments with dissimilar dynamics. In the special case of
identical environments i.e. when $\Big|\Big|{\bf P}_{i}-{\bf P}_{j}\Big|\Big|=0$,
the two bounds become the same.

We now prove an upper bound on the estimation error of the action-value
function of an optimal policy learned in $\boldsymbol{\mathrm{S}}_{j}$
when executed in the target environment $\mathcal{T}$.

\subsection*{Lemma 1}

Let $\pi_{1}^{*},...,\pi_{N}^{*}$ be $N$ optimal policies for ${\bf S}_{1},\ldots,{\bf S}_{N}$
respectively and $\tilde{Q}_{\mathcal{T}}^{\pi_{j}^{*}}=\big(\bm{\tilde{\psi}}^{\pi_{j}^{*}}\big)^{\mathrm{T}}\boldsymbol{\tilde{\mathrm{w}}}_{\mathcal{T}}$
denote the action-value function of an optimal policy learned in $\boldsymbol{\mathrm{S}}_{j}$
and executed in the target environment $\mathcal{T}$. Let $\bm{\tilde{\psi}}^{\pi_{j}^{*}}$
denote the estimated successor feature function from the combined
source and target observations from ${\bf S}_{j}$ and $\mathcal{T}$
as defined in Eq. (\ref{eq:MeanP}), and $\boldsymbol{\mathrm{\tilde{w}}}_{\mathcal{T}}$
is the estimated reward mapper for environment $\mathcal{T}$ by using
loss function in Eq. (\ref{eq:Loss}). It can be shown that the difference
of the true action-value function and the estimated one through successor
feature functions and reward mapper, is bounded as:
\[
\mathrm{Pr}\Big(\Big|{Q_{\mathcal{T}}}^{\pi_{j}^{*}}(s,a)-\tilde{Q}_{\mathcal{T}}^{\pi_{j}^{*}}(s,a)\Big|\leq\varepsilon(m)\ \forall s,a\Big)\geq1-\delta,
\]
where $\varepsilon(m)=\sqrt{2\mathrm{log}(|\mathbb{X}|u_{m}/\delta)}\sigma_{m,d}({\bf x}),\ {\bf x\in\mathbb{X}}\ \delta\in(0,1)$,
$u_{m}=\frac{\pi^{2}m^{2}}{6}$, $m$ being the number of observations
in environment $\mathcal{T}$, and ${\bf x}=(s,a)$. $\sigma_{m,d}({\bf x})$
is the square root of posterior variance as defined in Eq. (\ref{eq:GP}).

\textbf{Proof:} Proof is available in the supplementary material.

Lemma 1 ensures that reconstructing the action-value function on a
new target environment $\mathcal{T}$ by using $\bm{\tilde{\psi}}^{\pi_{j}^{*}}$
and $\boldsymbol{\mathrm{\tilde{w}}}_{\mathcal{T}}$ can be achieved
with a bounded error with high probability. Essentially, the key term
in $\varepsilon(m)$ is $\sigma_{m,d}^{2}({\bf x})$ that is computed
by using $\mathrm{\mathrm{\boldsymbol{\mathrm{K}}}_{*}}$ (see Eq.
(\ref{eq:KStar})), which itself incorporates the modeling noise variance
$\sigma_{{\bf S}}^{2}$. Hence, by increasing $\sigma_{{\bf S}}^{2}$,
$\sigma_{m,d}^{2}({\bf x})$ will be higher and accordingly the upper
bound will become looser. We note that due to the consistency of GPs,
as $m\rightarrow\infty$, the uncertainty of predictions tends to
0 ($\sigma_{m,d}^{2}({\bf x})\rightarrow0$) and thus $\varepsilon(m)\rightarrow0$.

We now present our final result that bounds the difference of the
optimal action-value function in the target environment and our GPI-derived
action value function.
\begin{figure*}[t]
\centering{}\includegraphics[scale=0.16]{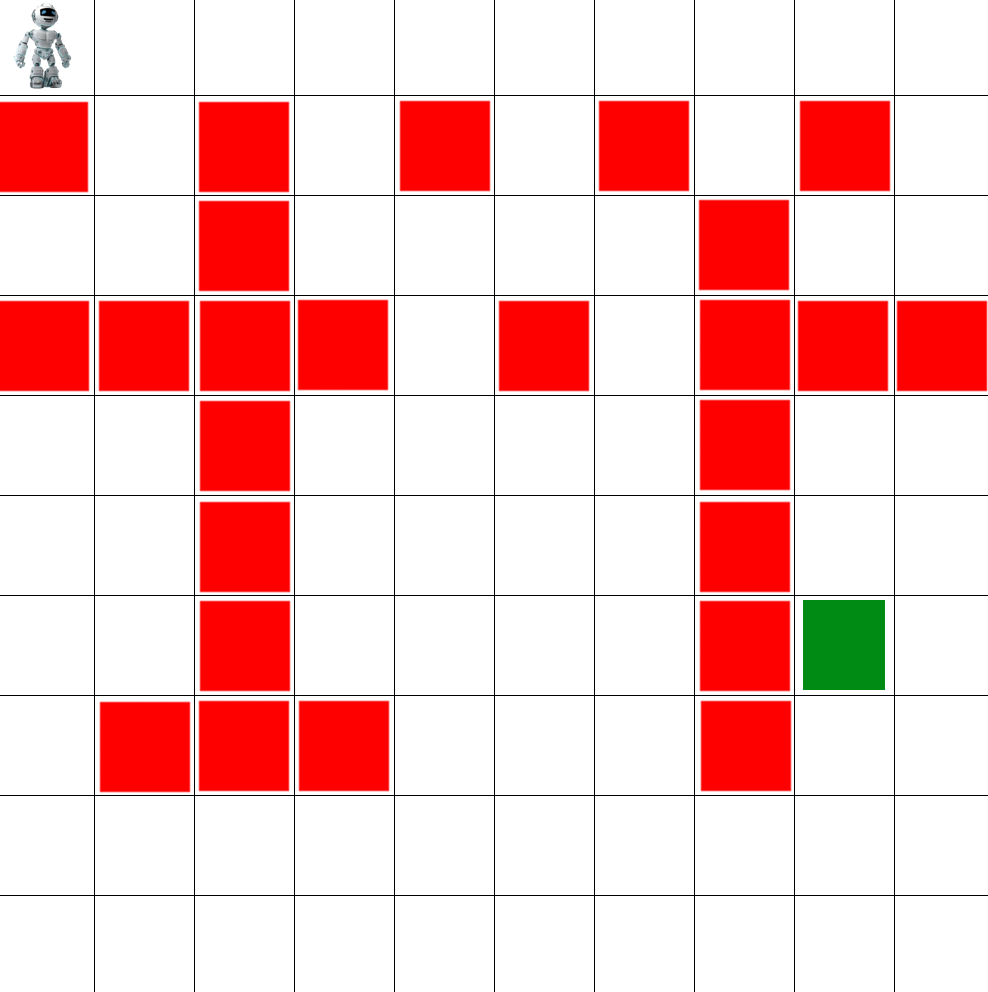}
\includegraphics[scale=0.45]{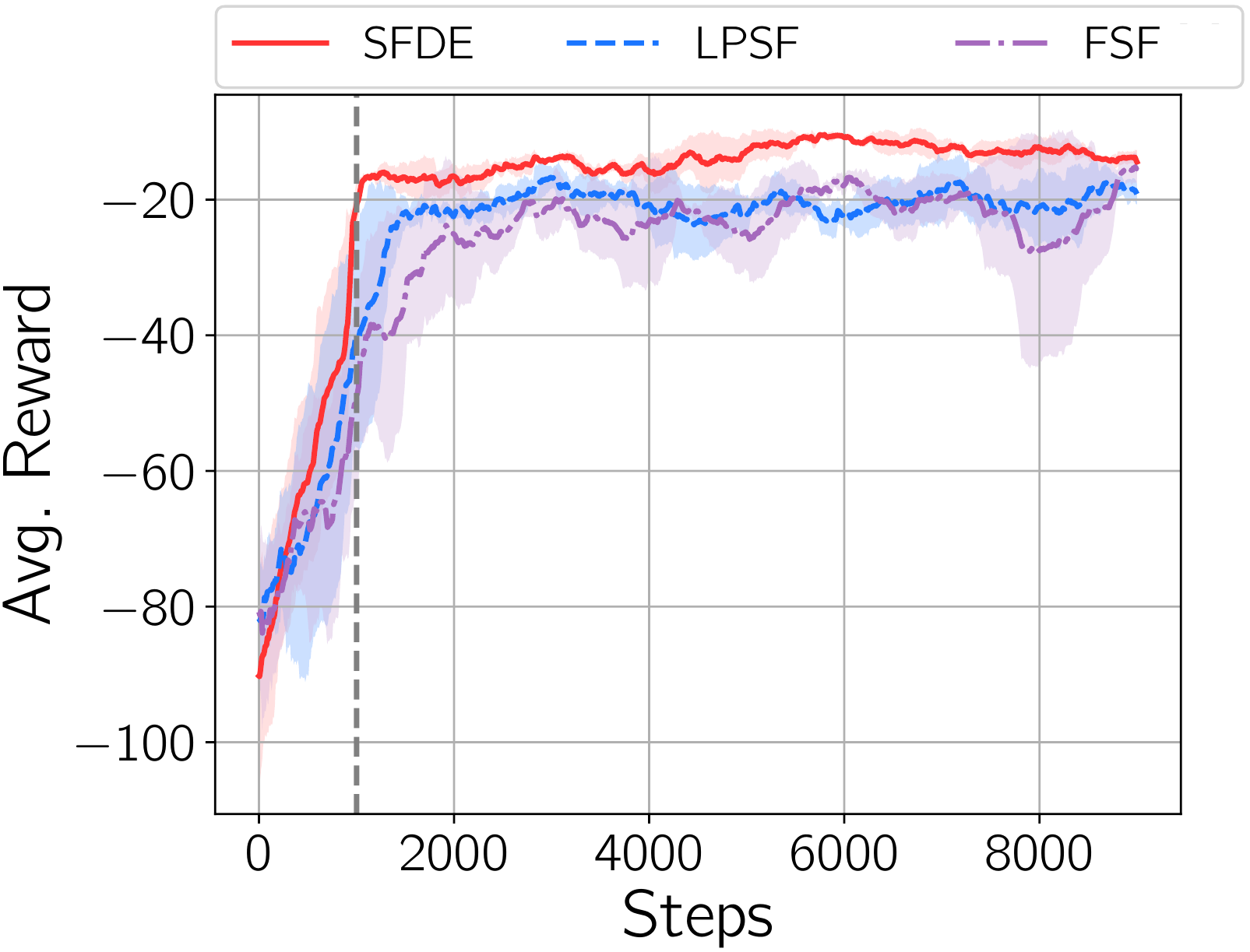}\caption{(left) Sample of proposed maze environment with red squares as obstacles
with -50 reward and green square as goal with +100 reward. (Right)
Obtained results of SFDE and two baselines. The results are averaged
over $50$ runs. The dashed vertical line demarcates the adaptation
phase.}
\label{fig:Exp_1}
\end{figure*}
\subsection*{Theorem 2 }

Let $\boldsymbol{\mathrm{S}}_{i=1...N}$ be $N$ different source
environments with dissimilar transition functions $p_{i=1...N}$.
Let us denote the optimal policy $\pi$ that is defined based on the
GPI as: 
\begin{equation}
\pi(s)\in\mathrm{\underset{a\in\mathcal{A}}{\argmaxx\ }\underset{j\in\{1...N\}}{\mathrm{max}}}\ \tilde{Q}_{\mathcal{T}}^{\pi_{j}^{*}}(s,a),\label{eq:t_eq_2}
\end{equation}
 where $\tilde{Q}_{\mathcal{T}}^{\pi_{j}^{*}}=\big(\bm{\tilde{\psi}}^{\pi_{j}^{*}}\big)^{\mathrm{T}}\boldsymbol{\tilde{\mathrm{w}}}_{\mathcal{T}}$
being the action-value function of an optimal policy learned in $\boldsymbol{\mathrm{S}}_{j}$
and executed in target environment $\mathcal{T}$, $\bm{\tilde{\psi}}^{\pi_{j}^{*}}$
is the estimated successor feature from the combined source and target
observations from ${\bf S}_{j}$ and $\mathcal{T}$ as defined in
Eq. (\ref{eq:MeanP}), and $\boldsymbol{\tilde{w}}_{\mathcal{T}}$
is the estimated reward mapper for target environment from Eq. (\ref{eq:Loss}).
Considering Lemma 1 and Eq. (\ref{eq:t_eq_2}), the difference of
optimal action-value function in the target environment and our GPI-derived
action value function is upper bounded as: 
\small
\begin{align}
Q_{\mathcal{T}}^{*}(s,a)-\tilde{Q}_{\mathcal{T}}^{\pi\in\pi_{j}^{*}}(s,a) & \leq\frac{2\phi_{\mathrm{max}}}{1-\gamma}\Big|\Big|{\bf \tilde{w}}_{\mathcal{T}}-{\bf w}_{j}\Big|\Big|\nonumber \\
+ & \frac{\gamma\Big|\Big|{\bf P}_{\mathcal{T}}(s,a)-{\bf P}_{j}(s,a)\Big|\Big|}{(1-\gamma)}\nonumber \\
\times & \frac{\Big(\Big|\Big|{\bf Q}_{\mathcal{T}}^{*}-{\bf Q}_{j}^{j}\Big|\Big|+\Big|\Big|{\bf Q}_{j}^{j}-{\bf Q}_{\mathcal{T}}^{j}\Big|\Big|\Big)}{(1-\gamma)}\nonumber \\
+ & \frac{2\varepsilon(m)}{(1-\gamma)}.\label{eq:proof2}
\end{align}
\normalsize

where $\phi_{\mathrm{max}}=\mathrm{max}_{s,a}||\phi(s,a)||$. We also
define ${\bf P}_{\mathcal{T}}=[p_{\mathcal{T}}(s^{\prime}|s,a),...]_{\forall s^{\prime}\in\mathcal{S}}$,
${\bf P}_{j}=[p_{j}(s^{\prime}|s,a),...]_{\forall s^{\prime}\in\mathcal{S}},$
${\bf Q}_{\mathcal{T}}^{*}=[\underset{b\in\mathcal{A}}{\mathrm{max}}{Q}_{\mathcal{T}}^{*}(s^{\prime},b),...]_{\forall s^{\prime}\in\mathcal{S}},$
${\bf Q}_{j}^{j}=[\underset{b\in\mathcal{A}}{\mathrm{max}}{\tilde{Q}}_{j}^{\pi_{j}^{*}}(s^{\prime},b),...]_{\forall s^{\prime}\in\mathcal{S}}$,
${\bf Q}_{\mathcal{T}}^{j}=[\underset{b\in\mathcal{A}}{\mathrm{max}}{\tilde{Q}}_{\mathcal{T}}^{\pi_{j}^{*}}(s^{\prime},b),...]_{\forall s^{\prime}\in\mathcal{S}}$,
and $\gamma$ as the discount factor.

\textbf{Proof:} Proof is available in supplementary material.

In inequality (\ref{eq:proof2}), $\Big|\Big|{\bf \tilde{w}}_{\mathcal{T}}-{\bf w}_{j}\Big|\Big|$
encodes the dissimilarity of reward functions in target environment
$\mathcal{T}$ and ${\bf S}_{j}$, and $\varepsilon(m)$ holds the
error of action-value reconstruction by GP-modelled successor features
(Lemma 1). Additionally, $\gamma||{\bf P}_{\mathcal{T}}-{\bf P}_{j}||\times(||{\bf Q}_{\mathcal{T}}^{*}-{\bf Q}_{j}^{j}||+||{\bf Q}_{j}^{j}-{\bf Q}_{\mathcal{T}}^{j}||)/(1-\gamma)$
is a term related to both the dissimilarity of dynamics in environment
$j\in\{1,\ldots.N\}$ and the target environment $\mathcal{T}$, as
well as the expected future reward in these two environments. We note
that Eq. (\ref{eq:t_eq_2}) enforces the selection of the best policy
among $j\in\{1,\ldots.N\}$ policies, hence the derived upper bound
is only based on the best selected policy among $N$ source environments. 

Theorem 2 is the core of our theoretical analysis that shows by using
$N$ different action-value functions that are obtained by their corresponding
successor feature functions and reward mapper weights, one can still
hold an upper bound on the difference of the optimal policy $Q_{\mathcal{T}}^{*}(s,a)$,
and $\tilde{Q}_{\mathcal{T}}^{\pi}(s,a)$, if $\pi(s)$ is selected
by GPI as defined in Eq. (\ref{eq:t_eq_2}). As expected, this upper
bound is looser when the norm distance between ${\bf \tilde{w}}_{\mathcal{T}}$
and ${\bf w}_{j}$ are higher - i.e. the reward functions are significantly
different. As explained in Lemma 1, by increasing the amount of noise
when modeling the source successor feature functions, $\sigma_{m,d}({\bf x})$
will be higher and accordingly the upper bound will become looser
since $\varepsilon(m)$ increases. This is reasonable as increasing
$\sigma_{\mathbf{S}}$ indicates that source and target environment
are significantly dissimilar. Note that if the environments are exactly
similar, $\gamma||{\bf P}_{\mathcal{T}}-{\bf P}_{j}||\times(||{\bf Q}_{\mathcal{T}}^{*}-{\bf Q}_{j}^{j}||+||{\bf Q}_{j}^{j}-{\bf Q}_{\mathcal{T}}^{j}||)/(1-\gamma)=0$,
this leads to a similar upper bound in \cite{barreto2018transfer}.
Further analysis on the obtained upper bound can be found in supplementary
materials. 

\section{Experiments}
We evaluate the performance of our method on 3 benchmarks: (1) A toy
navigation problem, (2) Classic CartPole control, and (3) The environment
introduced by Barreto et al. \cite{barreto2020fast,barreto2019option}.
We compare our approach, Successor Features for Dissimilar Environments
(\emph{SFDE}) with two related studies: Fast Successor Features (\emph{FSF})
\cite{barreto2020fast} and Linear Projection of Successor Features
(\emph{LPSF}) \cite{zhang2017deep}. Additional experiments are
available in the supplementary materials.
All the algorithms, including SFDE, are used in two phases: (1) The
first phase is adaptation where we fine-tune the source successor
feature functions using the first $1000$ target observations. This
step is method specific. (2) A testing phase in which we only use
the learnt policy and collect reward without updating the models.
For the adaptation phase, $\epsilon$-greedy based exploration is
used, afterwards we set the exploration rate $\varepsilon_{e}=0$
in the testing phase to demonstrate the effects of transfer from previous
environments. The baseline FSF is not equipped to handle environments
with dissimilar dynamics. We adapt this approach by fine-tuning the
successor feature functions of the source environments using the first
stored $1000$ observations of the target environment in the adaptation
phase. A batch size of $64$ is used at every step $64\leq t\leq1000$
to feed the new target observations to the previously learned source
successor features. Once the testing phase starts ($t>1000$), we
stop fine-tuning of the successor models for the rest of the experiment.
For the LPSF baseline, we follow the idea of \cite{zhang2017deep}
by defining a linear relation between the source and target successor
feature - that is there exists a mapping $\bm{\beta}$=$\bm{\beta}_{1}...\bm{\beta}_{N}$
such that the following loss function is minimised: $\mathcal{L}_{\bm{\beta}}(\bm{\theta}_{\beta})=\sum_{i=1...N}\Big|\Big|\tilde{\psi}^{\mathcal{T}}(\boldsymbol{\mathrm{x}})-\bm{\beta}_{i}\tilde{\psi}^{\boldsymbol{\mathrm{S}}_{i}}(\boldsymbol{\mathrm{x}})\Big|\Big|,\forall\boldsymbol{\mathrm{x}}\in\mathcal{D^{\bm{T}}}$.
Intuitively, the best linear projection is found for each source successor
feature function to minimise its distance to the target successor.
We follow the same approach explained for FSF to feed the target observations
in adaptation phase to $N$ neural networks that each represent the
model of $i-$th source successor features $i=\{1,\ldots,N\}$ with
$\mathcal{L}_{\bm{\beta}}$ loss function. Each of these $N$ neural
networks are MLPs with no hidden layers that is an equivalent of linear
regression in which the weights of the neural networks are $\bm{\beta}_{i}$.
Likewise, the best obtained value of $\bm{\beta}$ in the adaptation
phase is used in the testing phase. We used the same batch size of
$64$ to minimise $\mathcal{L}_{\bm{\beta}}$ loss function for LPSF.
The linearly projected successor features then used in the GPI framework
to obtain the optimal policy. In our approach, we construct the GP
based on the combination of source and target observations. To this
end, $500$ randomly sampled observations from source $\mathcal{D}^{\mathcal{\mathbf{S}}},\ \forall\mathcal{\mathbf{S}}\in\{{\bf S},\ldots,{\bf S}_{N}\}$
and all the observations in $\mathcal{D}^{\bm{\mathcal{T}}}$ at every
step of the adaptation phase is used. However, once the testing phase
is initiated, the new target observations are disregarded and previously
seen observations from the target are reused. Further details of implementations
are available in the supplementary material.

\subsection{Toy Navigation Problem}
The proposed maze problem consists of a $10\times10$ grid and the
agent needs to find the goal in the maze. The agent can pass through
the obstacles but it receives a reward of $-50$. It also receives
 $-1$ reward for each step and $+100$ reward for reaching the goal.
The action set is defined as $\mathcal{A}=\{\mathrm{left,right,up,down}\}$.
Figure \ref{fig:Exp_1} (left) shows an example of this environment
with red squares indicating obstacles, and the green square as the
goal. 

A set of $12$ policies $\Pi^{\mathrm{\mathbf{S}}}=\big\{\pi^{\mathrm{\mathbf{S}}_{1}},\ldots,\pi^{\mathrm{\mathbf{S}}_{12}}\big\}$
is learnt using generic Q-Learning on randomly generated maze source
environments $\Big\{\mathcal{M}(\mathcal{S},\mathcal{A},p_{1},R_{1}),\ldots,\mathcal{M}(\mathcal{S},\mathcal{A},p_{12},R_{12})\Big\}$
as ${\bf S}_{1},\ldots,{\bf S}_{12}$ in which the location of 25
obstacles and goal are changed. After obtaining the corresponding
policies, successor feature functions of these source environments
are learnt following Algorithm \ref{alg:Extracting-Successor-Features}.
Given $\bm{\tilde{\psi}}^{\pi_{1}},\ldots,\bm{\tilde{\psi}}^{\pi_{12}},$
we generate a random target environment and use the first $1000$
observations of adaptation phase as explained. Following Algorithm
\ref{alg:Successor-Features-for}, $\mathcal{GP}^{1...12}$ is constructed
as each step by fitting both source and target observations. We set
$\sigma_{{\bf S}}^{2}=0.1$ and $\sigma^{2}=0.01$ as the modelling
noise and the measurement noise, respectively. Having modelled the
successor features, we then perform GPI by using the predicted mean
as shown in Eq. (\ref{eq:MeanP}) and (\ref{eq:GP}). To calculate
$\boldsymbol{\tilde{w}}_{\mathcal{T}}$, the reward mapper of target
environment is calculated by minimising loss function introduced in
Eq. (\ref{eq:Loss}). Figure \ref{fig:Exp_1} (right) shows the performance
of our approach and other baselines. As expected, our method incorporating
the dissimilarity of environments performs better than the other two
related approaches. Figure \ref{fig:Exp_1} shows that LPSF adjusts
the source successor features but it seems to be slower than SFDE
in updating the successor feature functions as a linear projection
may not always be found. It can be seen that FSF can update the values
of successor features to some extent, however, it seems to be less
effective than the other 2 approaches. This can be the result of fine-tuning
the successor feature models with significantly different observations
that can be an issue in such scenarios. Note that we have not used
any visual information in this experiment and the location of agent
is translated to $(x,y)$ in the $10\times10$ grid.
\begin{figure}[t]
\begin{center}
\includegraphics[width=0.43\textwidth]{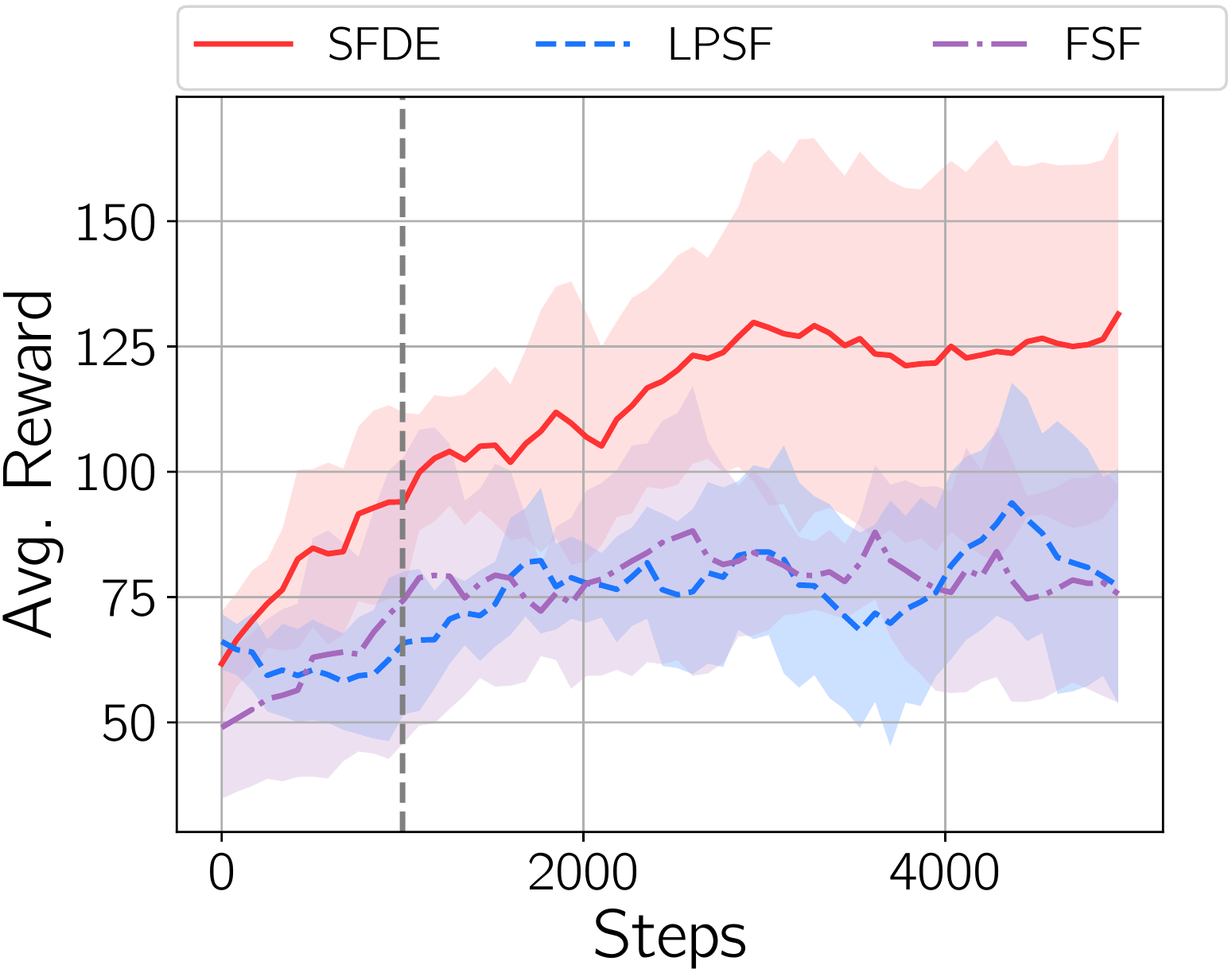}
\caption{Experiments with CartPole-v0 with $Pole\_Length=0.5m$ in source environment
and $Pole\_Length=3m$ in target environment. The results are averaged
over $10$ runs. The dashed vertical line demarcates the adaptation
phase.\label{fig:Experiments-with-CartPole-v0}}
\end{center}
\end{figure}
\subsection{CartPole-v0}
In the CartPole problem, we define a source environment ${\bf S}_{1}$
with a learnt policy $\pi^{{\bf S}_{1}}$ and a target environment
$\mathcal{T}.$ We incorporate the dissimilarities of dynamics in
source and target by changing the length of the pole from $Pole\_Length=0.5m$
in source, to $Pole\_Length=3.0m$ in the target environment. This
change impacts the transition probabilities of the target environment,
hence, it can be considered as a change in dynamics. Similar to the
previous experiment, we use the first $1000$ observations in all
three methods in a same manner. At each step, we fit $\mathcal{GP}^{1}$
by combination of source and target observations. We set $\sigma_{{\bf S}}^{2}=0.1$,
$\sigma^{2}=0.01$, and the maximum number of steps in the CartPole
is set to $200$ in each episode. To translate the image data into
states that can be used in our framework, we used the flattened output
of the last convolution layer as the state of the CartPole environment.
The detailed structure of the proposed network is available in supplementary
materials. Figure \ref{fig:Experiments-with-CartPole-v0} shows the
results of this experiment and it can be seen that our proposed method
outperforms both FSF and LPSF by using the GP-based modelled successor
features. 
\begin{figure}
\begin{center}
\includegraphics[scale=0.48]{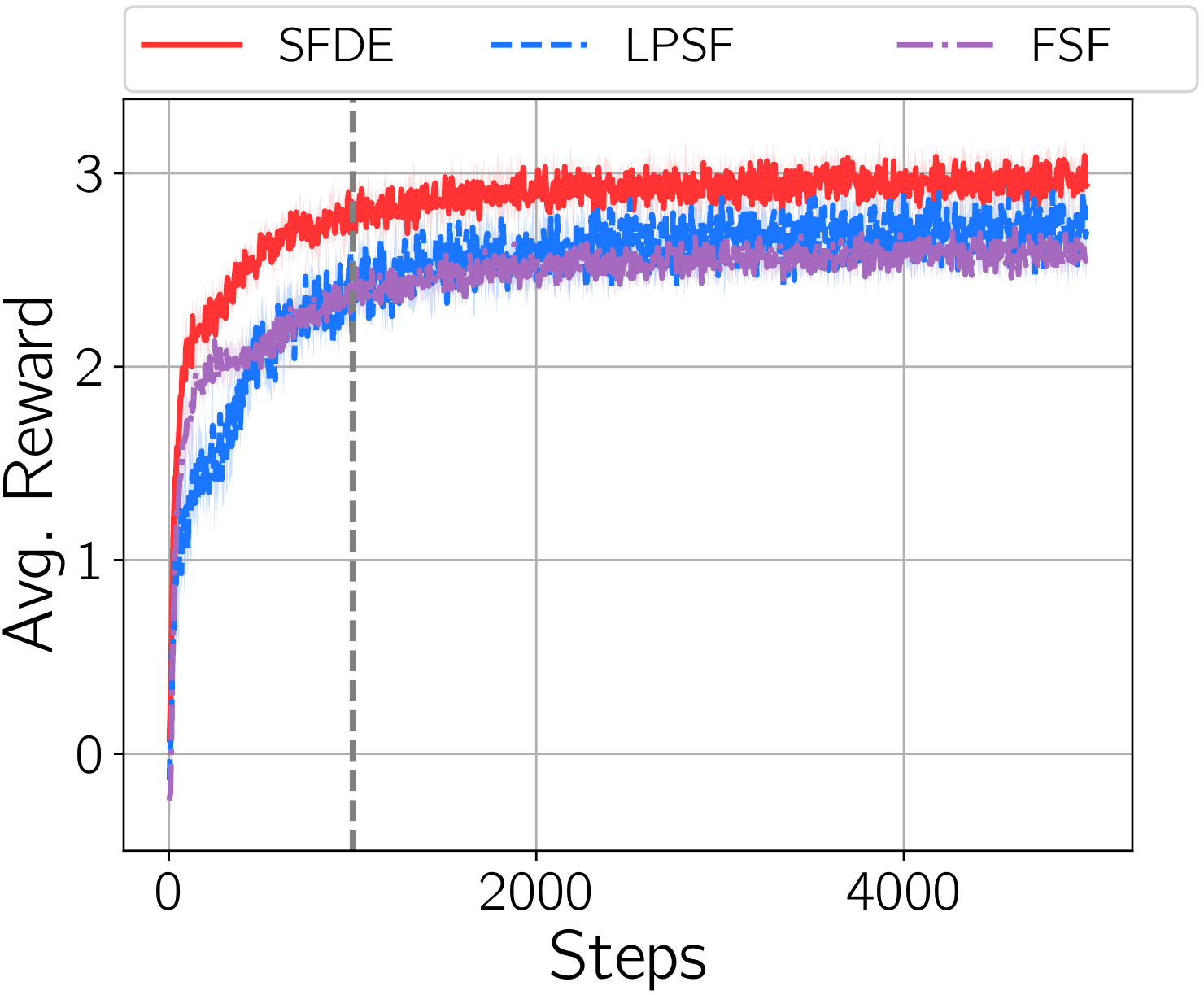}
\caption{Results on the environment introduced by FSF. The results are averaged
over $10$ runs. The dashed vertical line demarcates the adaptation
phase.\label{fig:Results-on-the}}
\end{center}%
\end{figure}

\subsection{FSF Environment}

We use the environment introduced in Barreto et al. \cite{barreto2020fast}
as our final experiment. The proposed environment is a $10\times10$
grid cells with $\mathcal{A}=\{\mathrm{left,right,up,down}\}$. There
are $10$ objects spread across the grid at all time that the agent
can pick up. Once an object is picked up, another random object will
appear in the grid randomly. Each object belongs to one of two types
(red or blue). At each time step, the agent receives an image showing
its position, the position of objects, and type of each object \cite{barreto2020fast}.
Then the agent selects the proper action to move in a direction. The
object is assumed picked up, if the agent occupies the cell in which
the object presents. In that case, it gets a reward based on the type
of the object and a new object will appear randomly in the grid. 

Following the setting of FSF experiments, we trained the agent in
$2$ different source environments with different reward functions
and dynamics of the environment. We incorporated the dissimilarity of
environments by adding a $5\%$ random transition noise to the target
environment and creating a single random terminal state with a negative
reward of $-1$. Figure \ref{fig:Results-on-the} shows the obtained
results on this problem. Similar to other experiments, our approach
seems to outperform the other two baselines. Interestingly, FSF outperforms
LPSF in adaptation phase, but both methods approximately converge
to the same value of the average reward.

\section{Conclusion}
In this paper we proposed a novel transfer learning approach based
on successor features in RL. Our approach is for the scenarios wherein
the source and the target environments have dissimilar reward functions
as well as dissimilar environment dynamics. We propose the use of
Gaussian Processes to model the source successor features functions
as noisy measurements of the target successor functions. We provide
a theoretical analysis on the convergence of our method by proving
an upper bound on the error of the optimal policy. We evaluate our
method on 3 benchmark problems and showed that our method outperform
existing methods.

\section*{Acknowledgements}
This research was partially funded by the Australian Government through
the Australian Research Council (ARC). Prof Venkatesh is the recipient
of an ARC Australian Laureate Fellowship (FL170100006).

\bibliography{example_paper}
\bibliographystyle{icml2021}
\clearpage

\section{Supplementary Materials}
\subsection{Theoretical Proofs}

We show our theoretical analysis for both cases of finite $\mathbb{X}$
and infinite $\mathbb{X}$.
\subsection{Finite $\mathbb{X}$}
\subsection*{Theorem 1}
Let ${\bf S}_{i}$ and ${\bf S}_{j}$ be two different source environments
with dissimilar transition dynamics $p_{i}$ and $p_{j}$ respectively.
Let $\delta_{ij}\triangleq\mathrm{max}_{s,a}\ |r_{i}(s,a)-r_{j}(s,a)|$,
where $r_{i}(.,.)$ and $r_{j}(.,.)$ are the reward functions of
environment ${\bf S}_{i}$ and ${\bf S}_{j}$ respectively. We denote
$\pi_{i}^{*}$ and $\pi_{j}^{*}$ as optimal policies in ${\bf S}_{i}$
and ${\bf S}_{j}$. It can be shown that the difference of their action-value
functions is upper bounded as:\small
\begin{align}
{Q}_{i}^{\pi_{i}^{*}}(s,a)-{Q}_{i}^{\pi_{j}^{*}}(s,a) & \leq\frac{2\delta_{ij}}{1-\gamma}\nonumber \\
+ & \frac{\gamma\Big|\Big|{\bf P}_{i}(s,a)-{\bf P}_{j}(s,a)\Big|\Big|}{(1-\gamma)}\nonumber \\
\times & \frac{\Big(\Big|\Big|{\bf Q}_{i}^{i}-{\bf Q}_{j}^{j}\Big|\Big|+\Big|\Big|{\bf Q}_{j}^{j}-{\bf Q}_{i}^{j}\Big|\Big|\Big)}{(1-\gamma)},\label{eq:eq:Inequality}
\end{align}
\normalsize
where ${Q}_{i}^{\pi_{k{\bf }}^{*}}$ shows the action-value function
in environment ${\bf S}_{i}$ by following an optimal policy that
is learned in the environment ${\bf S}_{k}\in\{{\bf S}_{1},\ldots,{\bf S}_{N}\}$.
We also define ${\bf P}_{i}(s,a)=[p_{i}(s^{\prime}|s,a),...]_{\forall s^{\prime}\in\mathcal{S}}$,
${\bf P}_{j}(s,a)=[p_{j}(s^{\prime}|s,a),...]_{\forall s^{\prime}\in\mathcal{S}},$
${\bf Q}_{i}^{i}=[\underset{b\in\mathcal{A}}{\mathrm{max}}{Q}_{i}^{\pi_{i}^{*}}(s^{\prime},b),...]_{\forall s^{\prime}\in\mathcal{S}},$
${\bf Q}_{j}^{j}=[\underset{b\in\mathcal{A}}{\mathrm{max}}{Q}_{j}^{\pi_{j}^{*}}(s^{\prime},b),...]_{\forall s^{\prime}\in\mathcal{S}}$,
${\bf Q}_{i}^{j}=[\underset{b\in\mathcal{A}}{\mathrm{max}}{Q}_{i}^{\pi_{j}^{*}}(s^{\prime},b),...]_{\forall s^{\prime}\in\mathcal{S}}$,
$\gamma$ as the discount factor, and $||.||$ to be $2-$norm (Euclidean
norm).
\textbf{Proof:}
We start by following the steps from \cite{barreto2018transfer}.
The left side of the inequality (\ref{eq:eq:Inequality}) can be rewritten as:
\begin{align*}
{Q}_{i}^{\pi_{i}^{*}}(s,a)-{Q}_{i}^{\pi_{j}^{*}}(s,a) & ={Q}_{i}^{\pi_{i}^{*}}(s,a)-{Q}_{j}^{\pi_{j}^{*}}(s,a)\\
+ & {Q}_{j}^{\pi_{j}^{*}}(s,a)-{Q}_{i}^{\pi_{j}^{*}}(s,a)\\
\leq & \underbrace{\Big|{Q}_{i}^{\pi_{i}^{*}}(s,a)-{Q}_{j}^{\pi_{j}^{*}}(s,a)\Big|}_{\text{(I)}}\\
+ & \underbrace{\Big|{Q}_{j}^{\pi_{j}^{*}}(s,a)-{Q}_{i}^{\pi_{j}^{*}}(s,a)\Big|}_{\text{(II)}}.
\end{align*}
For \textbf{(I)}, it can be shown that:\small
\begin{multline*}
\Big|{Q}_{i}^{\pi_{i}^{*}}(s,a)-{Q}_{j}^{\pi_{j}^{*}}(s,a)\Big|\leq\Big|r_{i}(s,a)+\gamma\sum_{s^{\prime}}p_{i}(s^{\prime}|s,a)\\
\underset{}{\mathrm{\underset{s^{\prime}}{\mathrm{max}}}\ {Q}_{i}^{\pi_{i}^{*}}(s^{\prime},b)}-r_{j}(s,a)-\gamma\sum_{s^{\prime}}p_{j}(s^{\prime}|s,a)\underset{}{\mathrm{\underset{b}{\mathrm{max}}}\ {Q}_{j}^{\pi_{j}^{*}}(s^{\prime},b)}\Big|\leq\\
\Big|r_{i}(s,a)-r_{j}(s,a)\Big|+\Big|\gamma\sum_{s^{\prime}}p_{i}(s^{\prime}|s,a)\underset{}{\mathrm{\underset{b}{\mathrm{max}}}\ {Q}_{i}^{\pi_{i}^{*}}(s^{\prime},b)}-\\
\gamma\sum_{s^{\prime}}p_{j}(s^{\prime}|s,a)\underset{}{\mathrm{\underset{b}{\mathrm{max}}}\ {Q}_{j}^{\pi_{j}^{*}}(s^{\prime},b)}\Big|\leq\\
\Big|r_{i}(s,a)-r_{j}(s,a)\Big|+\gamma\Big(\Big|\Big|\big({\bf P}_{i}-{\bf P}_{j}\big).\big({\bf Q}_{i}^{i}-{\bf Q}_{j}^{j}\big)\Big|\Big|\Big)\leq\\
\Big|r_{i}(s,a)-r_{j}(s,a)\Big|+\gamma\Big|\Big|{\bf P}_{i}-{\bf P}_{j}\Big|\Big|\times\Big|\Big|{\bf Q}_{i}^{i}-{\bf Q}_{j}^{j}\Big|\Big|=\\
\delta_{ij}+\gamma\Big|\Big|{\bf P}_{i}-{\bf P}_{j}\Big|\Big|\times\Big|\Big|{\bf Q}_{i}^{i}-{\bf Q}_{j}^{j}\Big|\Big|\leq\\
\frac{\delta_{ij}}{1-\gamma}+\frac{\gamma}{1-\gamma}\Big|\Big|{\bf P}_{i}-{\bf P}_{j}\Big|\Big|\times\Big|\Big|{\bf Q}_{i}^{i}-{\bf Q}_{j}^{j}\Big|\Big|,\ \forall s^{\prime}\in\mathcal{S}
\end{multline*}
\normalsize
For \textbf{(II)}, it can be shown that:
\small
\begin{multline*}
\Big|{Q}_{j}^{\pi_{j}^{*}}(s,a)-{Q}_{i}^{\pi_{j}^{*}}(s,a)\Big|\leq\Big|r_{j}(s,a)+\gamma\sum_{s^{\prime}}p_{j}(s^{\prime}|s,a)\\
\underset{}{\mathrm{\underset{b}{\mathrm{max}}}\ {Q}_{j}^{\pi_{j}^{*}}(s^{\prime},b)}-r_{i}(s,a)-\gamma\sum_{s^{\prime}}p_{i}(s^{\prime}|s,a)\underset{}{\mathrm{\underset{b}{\mathrm{max}}}\ {Q}_{i}^{\pi_{j}^{*}}(s^{\prime},b)}\Big|\leq\\
\Big|r_{j}(s,a)-r_{i}(s,a)\Big|+\Big|\gamma\sum_{s^{\prime}}p_{j}(s^{\prime}|s,a)\underset{}{\mathrm{\underset{b}{\mathrm{max}}}\ {Q}_{j}^{\pi_{j}^{*}}(s^{\prime},b)}-\\
\gamma\sum_{s^{\prime}}p_{i}(s^{\prime}|s,a)\underset{}{\mathrm{\underset{b}{\mathrm{max}}}\ {Q}_{i}^{\pi_{j}^{*}}(s^{\prime},b)}\Big|\leq\\
\Big|r_{j}(s,a)-r_{i}(s,a)\Big|+\gamma\Big(\Big|\Big|\big({\bf P}_{j}-{\bf P}_{i}\big).\big({\bf Q}_{j}^{j}-{\bf Q}_{i}^{j}\big)\Big|\Big|\Big)\leq\\
\Big|r_{j}(s,a)-r_{i}(s,a)\Big|+\gamma\Big|\Big|{\bf P}_{j}-{\bf P}_{i}\Big|\Big|\times\Big|\Big|{\bf Q}_{j}^{j}-{\bf Q}_{i}^{j}\Big|\Big|=\\
\delta_{ij}+\gamma\Big|\Big|{\bf P}_{j}-{\bf P}_{i}\Big|\Big|\times\Big|\Big|{\bf Q}_{j}^{j}-{\bf Q}_{i}^{j}\Big|\Big|\leq\\
\frac{\delta_{ij}}{1-\gamma}+\frac{\gamma}{1-\gamma}\Big|\Big|{\bf P}_{j}-{\bf P}_{i}\Big|\Big|\times\Big|\Big|{\bf Q}_{j}^{j}-{\bf Q}_{i}^{j}\Big|\Big|=\\
\frac{\delta_{ij}}{1-\gamma}+\frac{\gamma}{1-\gamma}\Big|\Big|{\bf P}_{i}-{\bf P}_{j}\Big|\Big|\times\Big|\Big|{\bf Q}_{j}^{j}-{\bf Q}_{i}^{j}\Big|\Big|,\ \forall s^{\prime}\in\mathcal{S}
\end{multline*}
\normalsize
Considering (I) and (II), it leads to the upper bound:
\begin{align*}
{Q}_{i}^{\pi_{i}^{*}}(s,a)-{Q}_{i}^{\pi_{j}^{*}}(s,a) & \leq\frac{2\delta_{ij}}{1-\gamma}+\frac{\gamma\Big|\Big|{\bf P}_{i}-{\bf P}_{j}\Big|\Big|}{(1-\gamma)}\\
\times & \frac{\Big(\Big|\Big|{\bf Q}_{i}^{i}-{\bf Q}_{j}^{j}\Big|\Big|+\Big|\Big|{\bf Q}_{j}^{j}-{\bf Q}_{i}^{j}\Big|\Big|\Big)}{(1-\gamma)}.\\
 & \ \ \ \ \!\ \ \ \ \ \ \ \!\ \ \ \ \ \ \ \ \ \ \ \ \ \ \ \ \ \ \ \ \ \ \ \ \ \ \ \ \spadesuit
\end{align*}

\subsection*{Lemma 1}
Let $\pi_{1}^{*},...,\pi_{N}^{*}$ be $N$ optimal policies for ${\bf S}_{1},\ldots,{\bf S}_{N}$
respectively and $\tilde{Q}_{\mathcal{T}}^{\pi_{j}^{*}}=\big(\bm{\tilde{\psi}}^{\pi_{j}^{*}}\big)^{\mathrm{T}}\boldsymbol{\tilde{\mathrm{w}}}_{\mathcal{T}}$
denote the action-value function of an optimal policy learned in $\boldsymbol{\mathrm{S}}_{j}$
and executed in the target environment $\mathcal{T}$. Let $\bm{\tilde{\psi}}^{\pi_{j}^{*}}$
denote the estimated successor feature function from the combined
source and target observations from ${\bf S}_{j}$ and $\mathcal{T}$
as defined in Eq. (9)(in paper), and $\boldsymbol{\mathrm{\tilde{w}}}_{\mathcal{T}}$
is the estimated reward mapper for environment $\mathcal{T}$ by using
loss function in Eq. (6)(in paper). It can be shown that the difference
of the true action-value function and the estimated one through successor
feature functions and reward mapper, is bounded as:
\[
\mathrm{Pr}\Big(\Big|{Q_{\mathcal{T}}}^{\pi_{j}^{*}}(s,a)-\tilde{Q}_{\mathcal{T}}^{\pi_{j}^{*}}(s,a)\Big|\leq\varepsilon(m)\ \forall s,a\Big)\geq1-\delta,
\]
where $\varepsilon(m)=\sqrt{2\mathrm{log}(|\mathbb{X}|u_{m}/\delta)}\sigma_{m,d}({\bf x}),\ {\bf x\in\mathbb{X}}\ \delta\in(0,1)$,
$u_{m}=\frac{\pi^{2}m^{2}}{6}$, $m$ being the number of observations
in environment $\mathcal{T}$, and ${\bf x}=(s,a)$. $\sigma_{m,d}({\bf x})$
is the square root of posterior variance as defined in Eq. (10)(in
paper).

\textbf{Proof:} 
For proving this Lemma, we first follow the properties of Normal distribution.
Let us assume that $l\sim\mathcal{N}(0,1),$ and ${\psi_{\mathcal{T},d}^{\pi_{j}^{*}}}({\bf x})\sim\mathcal{N}\big(\mu_{m,d}({\bf x}),\sigma_{m,d}^{2}({\bf x})\big)$,
${\bf x}\in\mathbb{X},$ $\mathrm{{\bf x}}=(s,a)$, and $d=\{1,\ldots,D\}$
as defined in Eq. (9) and Eq. (10)(in paper). $m$ target observations
are assumed to be available. We follow \cite{srinivas2009gaussian},
based on the properties of Normal distribution, $\mathrm{Pr}(l>c),\ c>0$
is calculated as:
\begin{align*}
\mathrm{Pr}(l & >c)=e^{-c^{2}/2}(2\pi)^{-1/2}\int e^{-(l-c)^{2}-c(l-c)}dl\leq\\
e^{-c^{2}/2} & \mathrm{Pr}(l>0)=\frac{1}{2}(e^{-c^{2}/2}).
\end{align*}

As $c>0$, we know $e^{-c(l-c)}\leq1$ for $l\geq c.$ Accordingly,
$\mathrm{Pr}\Big[|{\psi_{\mathcal{T},d}^{\pi_{j}^{*}}}(\mathrm{{\bf x}})-\tilde{\psi}_{\mathcal{T},d}^{\pi_{j}^{*}}(\mathrm{{\bf x}})|>\beta_{m}^{1/2}\sigma_{m,d}(\mathrm{{\bf x}})\Big]\leq e^{-\beta_{m}/2},\ \beta_{m}=2\mathrm{log}(|\mathbb{X}|u_{m}/\delta).$
Now assuming $l=\big({\psi_{\mathcal{T},d}^{\pi_{j}^{*}}}(\mathrm{{\bf x}})-\tilde{\psi}_{\mathcal{T},d}^{\pi_{j}^{*}}(\mathrm{{\bf x}})\big)/\sigma_{m,d}(\mathrm{{\bf x}})$
and $c=\beta_{m}^{1/2}$, the error of modelling successor feature
function can be written as:
\begin{align}
\mathrm{Pr}\Big(\Big|{\psi_{\mathcal{T},d}^{\pi_{j}^{*}}}(\mathrm{{\bf x}})-\tilde{\psi}_{\mathcal{T},d}^{\pi_{j}^{*}}(s,a)\Big| & \leq\beta_{m}^{1/2}\sigma_{m,d}(\mathrm{{\bf x}})\Big)\nonumber \\
\geq & 1-|\mathbb{X}|e^{-\beta_{m}/2}.\ \ \forall\mathrm{\mathrm{{\bf x}}}\in\mathbb{X}.\label{eq:PrLemma}
\end{align}

If $|\mathbb{X}|e^{-\beta_{m}/2}=\frac{\delta}{u_{m}}$, the inequality
\ref{eq:PrLemma} holds for $u_{m}=\pi^{2}m^{2}/6$. We follow the
assumption in \cite{barreto2017successor,barreto2019option,barreto2020fast},
$\exists\boldsymbol{\tilde{\mathrm{w}}}_{\mathcal{T}},\ s.t.\ \tilde{Q}_{\mathcal{T}}^{\pi_{j}^{*}}=\big(\bm{\tilde{\psi}}^{\pi_{j}^{*}}\big)^{\mathrm{T}}\boldsymbol{\tilde{\mathrm{w}}}_{\mathcal{T}}$
given all dimensions of successor feature function, hence Lemma 1
holds for $\forall\mathrm{{\bf x}}\in\mathbb{X}:$
\begin{align*}
\mathrm{Pr}\Big(\Big|{Q_{\mathcal{T}}}^{\pi_{j}^{*}}(\mathrm{{\bf x}})-\tilde{Q}_{\mathcal{T}}^{\pi_{j}^{*}}(\mathrm{{\bf x}})\Big| & \leq\varepsilon(m)\ \forall\mathrm{{\bf x}}\in\mathbb{X}\Big)\geq1-\delta.\\
 & \ \ \ \ \!\ \ \ \ \ \ \ \!\ \ \ \ \ \ \ \ \ \ \spadesuit
\end{align*}
\\
We note $\varepsilon(m)$ decreases $\varepsilon(m)\in\big(\mathcal{O}\mathrm{log}(m)^{-1}\big)$
as $\sigma_{m,d}({\bf x})\in\mathcal{O}\big(\mathrm{log}(m)^{-2}\big)$
\cite{lederer2019uniform} and $\sqrt{2\mathrm{log}(|\mathbb{X}|u_{m}/\delta)}\in\mathcal{O}\big(\mathrm{log}(m)\big)$.
This guarantees that the error of modelling convergence to zero as
$m\rightarrow\infty$. Before starting the proof of Theorem 2, we
present Remark 1 based on the concept of GPI \cite{barreto2018transfer}
as follows:

\paragraph*{Remark 1 }
Let $\pi_{1},...,\pi_{N}$ be $N$ decision policies and correspondingly
$\tilde{Q}^{\pi_{1}},\tilde{Q}^{\pi_{2}},...,\tilde{Q}^{\pi_{N}}$
are the respective estimated action-value functions (Lemma 1) such
that: 
\[
\Big|{Q}^{\pi_{i}}(\boldsymbol{\mathrm{x}})-\tilde{Q}^{\pi_{i}}(\boldsymbol{\mathrm{x}})\Big|\leq\varepsilon(m)\ \ \forall\boldsymbol{\mathrm{x}}=(s,a),
\]
 where $m$ is the number of target observations. Defining: $\pi(s)\in\mathrm{argmax_{a}\ max_{i}}\ \tilde{Q}^{\pi_{i}}(s,a)$,
Then: 
\[
{Q}^{\pi}(\boldsymbol{\mathrm{x}})\geq\underset{i}{\mathrm{max}}\ {Q}^{\pi_{i}}(\boldsymbol{\mathrm{x}})-\frac{2}{1-\gamma}\varepsilon(m)\ \ \forall\boldsymbol{\mathrm{x}}=(s,a).
\]

\textbf{Proof:}

We start the proof by extending $P^{\pi}\mathrm{\underset{i}{\mathrm{max}}}\ {\tilde{Q}}^{\pi_{i}}(s,a)$,
where $P^{\pi}$ is the Bellman operator \cite{barreto2018transfer}:
\small
\begin{multline*}
P^{\pi}\mathrm{\underset{i}{\mathrm{max}}}\ {\tilde{Q}}^{\pi_{i}}(s,a)=r(s,a)+\gamma\sum_{s^{\prime}}p_{i}(s^{\prime}|s,a)\mathrm{\underset{i}{\mathrm{max}}}\ {\tilde{Q}}^{\pi_{i}}(s^{\prime},\pi(s^{\prime}))\\
\geq r(s,a)+\gamma\sum_{s^{\prime}}p_{i}(s^{\prime}|s,a)\mathrm{\underset{b}{\mathrm{max}}}\ {Q}^{\pi_{i}}(s^{\prime},b)-\gamma\varepsilon(m)\\
\geq r(s,a)+\gamma\sum_{s^{\prime}}p_{i}(s^{\prime}|s,a)\mathrm{\underset{i}{\mathrm{max}}}\ {Q}^{\pi_{i}}(s^{\prime},\pi_{i}(s^{\prime}))-\gamma\varepsilon(m)\\
\geq r(s,a)+\gamma\sum_{s^{\prime}}p_{i}(s^{\prime}|s,a)\ {Q}^{\pi_{i}}(s^{\prime},\pi_{i}(s^{\prime}))-\gamma\varepsilon(m)\\
={Q}^{\pi_{i}}(s,a)-\gamma\varepsilon(m)\\
\end{multline*}

as $P^{\pi}\mathrm{\underset{i}{\mathrm{max}}}\ {\tilde{Q}}^{\pi_{i}}(s,a)\geq{Q}^{\pi_{i}}(s,a)-\gamma\varepsilon(m),\ \forall i=\{1,\ldots,N\}$:
\begin{multline*}
P^{\pi}\mathrm{\underset{i}{\mathrm{max}}}\ {\tilde{Q}}^{\pi_{i}}(s,a)\geq\mathrm{\underset{i}{\mathrm{max}}}\ {Q}^{\pi_{i}}(s,a)-\gamma\varepsilon(m)\\
\geq{\tilde{Q}}^{\pi_{i}}(s,a)-\gamma\varepsilon(m)-\varepsilon(m)\\
\geq{\tilde{Q}}^{\pi_{i}}(s,a)-\frac{1+\gamma}{1-\gamma}\varepsilon(m)\\
\geq{Q}^{\pi_{i}}(s,a)-\varepsilon(m)-\frac{1+\gamma}{1-\gamma}\varepsilon(m)
\end{multline*}
\normalsize
We also know that $Q^{\pi}(\boldsymbol{\mathrm{x}})=\mathrm{lim}_{k\rightarrow\infty}(P^{\pi})^{k}\mathrm{\underset{i}{\mathrm{max}}}\ {\tilde{Q}}^{\pi_{i}}(\boldsymbol{\mathrm{x}})$.
Then, it follows:
\begin{multline*}
{Q}^{\pi}(\boldsymbol{\mathrm{x}})\geq\mathrm{\underset{i}{\mathrm{max}}}\ {Q}^{\pi_{i}}(\boldsymbol{\mathrm{x}})-\frac{2}{1-\gamma}\varepsilon(m)\ \ \forall\boldsymbol{\mathrm{x}}=(s,a),\boldsymbol{\mathrm{x}}\in\mathbb{X}.\\
\spadesuit
\end{multline*}

\subsection*{Theorem 2 }
Let $\boldsymbol{\mathrm{S}}_{i=1...N}$ be $N$ different source
environments with dissimilar transition functions $p_{i=1...N}$.
Let us denote the optimal policy $\pi$ that is defined based on the
GPI as: 
\begin{equation}
\pi(s)\in\mathrm{\underset{a\in\mathcal{A}}{\argmaxx\ }\underset{j\in\{1...N\}}{\mathrm{max}}}\ \tilde{Q}_{\mathcal{T}}^{\pi_{j}^{*}}(s,a),\label{eq:t_eq_2}
\end{equation}
 where $\tilde{Q}_{\mathcal{T}}^{\pi_{j}^{*}}=\big(\bm{\tilde{\psi}}^{\pi_{j}^{*}}\big)^{\mathrm{T}}\boldsymbol{\tilde{\mathrm{w}}}_{\mathcal{T}}$
being the action-value function of an optimal policy learned in $\boldsymbol{\mathrm{S}}_{j}$
and executed in target environment $\mathcal{T}$, $\bm{\tilde{\psi}}^{\pi_{j}^{*}}$
is the estimated successor feature from the combined source and target
observations from ${\bf S}_{j}$ and $\mathcal{T}$ as defined in
Eq. (9)(in paper), and $\boldsymbol{\tilde{w}}_{\mathcal{T}}$ is
the estimated reward mapper for target environment from Eq. (6)(in
paper). Considering Lemma 1 and Eq. (13)(in paper), the difference
of optimal action-value function in the target environment and our
GPI-derived action value function is upper bounded as: 
\small
\begin{align}
Q_{\mathcal{T}}^{*}(s,a)-\tilde{Q}_{\mathcal{T}}^{\pi\in\pi_{j}^{*}}(s,a) & \leq\frac{2\phi_{\mathrm{max}}}{1-\gamma}\Big|\Big|{\bf \tilde{w}}_{\mathcal{T}}-{\bf w}_{j}\Big|\Big|\nonumber \\
+ & \frac{\gamma\Big|\Big|{\bf P}_{\mathcal{T}}(s,a)-{\bf P}_{j}(s,a)\Big|\Big|}{(1-\gamma)}\nonumber \\
\times & \frac{\Big(\Big|\Big|{\bf Q}_{\mathcal{T}}^{*}-{\bf Q}_{j}^{j}\Big|\Big|+\Big|\Big|{\bf Q}_{j}^{j}-{\bf Q}_{\mathcal{T}}^{j}\Big|\Big|\Big)}{(1-\gamma)}\nonumber \\
+ & \frac{2\varepsilon(m)}{(1-\gamma)}.\label{eq:proof2}
\end{align}
\normalsize
where $\phi_{\mathrm{max}}=\mathrm{max}_{s,a}||\phi(s,a)||$. We also
define ${\bf P}_{\mathcal{T}}=[p_{\mathcal{T}}(s^{\prime}|s,a),...]_{\forall s^{\prime}\in\mathcal{S}}$,
${\bf P}_{j}=[p_{j}(s^{\prime}|s,a),...]_{\forall s^{\prime}\in\mathcal{S}},$
${\bf Q}_{\mathcal{T}}^{*}=[\underset{b\in\mathcal{A}}{\mathrm{max}}{Q}_{\mathcal{T}}^{*}(s^{\prime},b),...]_{\forall s^{\prime}\in\mathcal{S}},$
${\bf Q}_{j}^{j}=[\underset{b\in\mathcal{A}}{\mathrm{max}}{\tilde{Q}}_{j}^{\pi_{j}^{*}}(s^{\prime},b),...]_{\forall s^{\prime}\in\mathcal{S}}$,
${\bf Q}_{\mathcal{T}}^{j}=[\underset{b\in\mathcal{A}}{\mathrm{max}}{\tilde{Q}}_{\mathcal{T}}^{\pi_{j}^{*}}(s^{\prime},b),...]_{\forall s^{\prime}\in\mathcal{S}}$,
and $\gamma$ as the discount factor.

\textbf{Proof:}
$Q_{\mathcal{T}}^{*}(s,a)-\tilde{Q}_{\mathcal{T}}^{\pi\in\pi_{j}^{*}}(s,a)$
is defined as the difference of the optimal action-value function,
and the action-value function derived from GPI. It can be shown that:
\begin{multline*}
Q_{\mathcal{T}}^{*}(s,a)-\tilde{Q}_{\mathcal{T}}^{\pi}(s,a)\leq Q_{\mathcal{T}}^{*}(s,a)-Q_{\mathcal{T}}^{\pi_{j}^{*}}(s,a)+\frac{2}{1-\gamma}\varepsilon(m)\\
\leq\frac{2\delta_{ij}}{1-\gamma}+\frac{\gamma\Big|\Big|{\bf P}_{i}-{\bf P}_{j}\Big|\Big|\times\Big|\Big|{\bf Q}_{\mathcal{T}}^{*}-{\bf Q}_{j}^{j}\Big|\Big|+\Big|\Big|{\bf Q}_{j}^{j}-{\bf Q}_{\mathcal{T}}^{j}\Big|\Big|}{(1-\gamma)}\\
+\frac{2}{1-\gamma}\varepsilon(m)\mathit{\mathrm{\ \ \ \ \ //Theorem}}1\\
\leq\frac{2}{1-\gamma}\underset{s,a}{\mathrm{max}}||\phi(s,a)||\ ||\mathrm{\boldsymbol{w}_{\mathcal{T}}}-\boldsymbol{\mathrm{w}}_{j}||\\
+\frac{\gamma\Big|\Big|{\bf P}_{i}-{\bf P}_{j}\Big|\Big|\times\Big|\Big|{\bf Q}_{\mathcal{T}}^{*}-{\bf Q}_{j}^{j}\Big|\Big|+\Big|\Big|{\bf Q}_{j}^{j}-{\bf Q}_{\mathcal{T}}^{j}\Big|\Big|}{(1-\gamma)}+\frac{2}{1-\gamma}\varepsilon(m).\\
\\
\spadesuit
\end{multline*}

As explained in Lemma 1, $\varepsilon(m)\rightarrow0$ with $\varepsilon(m)\in\big(\mathcal{O}\mathrm{log}(m)^{-1}\big)$,
as $m\rightarrow\infty$ - that is, the number of target observations
tend to infinity. Note that the remaining terms of the upper bound
depends on the amount of dissimilarity of source and target environments
as explained in Section 3.1 of the paper.

\subsection{Infinite $\mathbb{X}$\label{subsec:Infinite}}
We now continue our analysis on the cases that the action-state space
($\mathbb{X}$) is infinite - i.e. there may be infinite observations
coming from the target environment. In that case, Lemma 1 will not
hold and further steps need to be taken.

Let us assume $\mathbb{X}_{m}\subset\mathbb{X}$ represents a subset
of infinite $\mathbb{X}$ at time step $m$, where $m$ target observations
are seen. Clearly, Lemma 1 will hold with this assumption if $\beta_{m}=2\mathrm{log}(|\mathbb{X}_{m}|u_{m}/\delta).$
The main question in here is if we can extend this to the whole search
space $\mathbb{X}.$ 

Following Boole's inequality - known as union bound, it can be shown
that for some constants $a,b,L>0$ \cite{srinivas2009gaussian}:
\begin{multline*}
\mathrm{Pr\Big(}\forall i=\{1,2\},\forall\boldsymbol{\mathrm{x}}\mathrm{\in\mathbb{X}},|\frac{\partial\psi_{\mathcal{T},d}^{\pi_{j}^{*}}}{\partial\boldsymbol{\mathrm{x}}_{i}}|<L\Big)\geq1-2ab^{\frac{L^{2}}{b^{2}}},
\end{multline*}

that implies:
\small
\begin{multline}
\mathrm{\Big(}\forall\mathrm{\boldsymbol{\mathrm{x}}\in\mathbb{X}},|\psi_{\mathcal{T},d}^{\pi_{j}^{*}}(\mathrm{\boldsymbol{\mathrm{x}}})-\psi_{\mathcal{T},d}^{\pi_{j}^{*}}(\mathrm{\boldsymbol{\mathrm{x}}^{\prime}})|\Big)\leq L|\mathrm{\boldsymbol{\mathrm{x}}-\mathrm{\boldsymbol{\mathrm{x}}^{\prime}}}|,\ \ \mathrm{\boldsymbol{\mathrm{x}},\mathrm{\boldsymbol{\mathrm{x}}^{\prime}}}\in\mathbb{X}.\label{eq:Dis}
\end{multline}
\normalsize
Eq. (\ref{eq:Dis}) enables us to perform a discretisation on the
search space $\mathbb{X}_{m}$ with size of $\tau_{m}^{2}$ so that:
\begin{multline*}
|\mathrm{\boldsymbol{\mathrm{x}}-\mathrm{[\boldsymbol{\mathrm{x}}]_{m}|\leq2r/}}\tau_{m},
\end{multline*}
where $[\mathrm{\boldsymbol{\mathrm{x}}}]_{m}$ denotes the closest
point from $\mathbb{X}_{m}$ to $\boldsymbol{\mathrm{x}}\in\mathbb{X}$
and $\tau_{m}$ implies the number uniformly spaced points on both
coordinates of $\mathbb{X}_{m}$ that is a discretisation factor.
We now proceed to Lemma 2 as an extension of successor feature function
modelling error in infinite $\mathbb{X}$ space. 

\subsection*{Lemma 2}
Let $\boldsymbol{\mathrm{x}}=(s,a)\in\mathbb{X}$, and $\mathbb{X}$
is infinite state-action space. $\pi_{1}^{*},...,\pi_{N}^{*}$ is
$N$ optimal policies for ${\bf S}_{1},\ldots,{\bf S}_{N}$ respectively
and $\tilde{Q}_{\mathcal{T}}^{\pi_{j}^{*}}=\big(\bm{\tilde{\psi}}^{\pi_{j}^{*}}\big)^{\mathrm{T}}\boldsymbol{\tilde{\mathrm{w}}}_{\mathcal{T}}$
denote the action-value function of an optimal policy learned in $\boldsymbol{\mathrm{S}}_{j}$
and executed in the target environment $\mathcal{T}$. Let $\bm{\tilde{\psi}}^{\pi_{j}^{*}}$
denote the estimated successor feature function from the combined
source and target observations from ${\bf S}_{j}$ and $\mathcal{T}$
as defined in Eq. (9)(in paper), and $\boldsymbol{\mathrm{\tilde{w}}}_{\mathcal{T}}$
is the estimated reward mapper for environment $\mathcal{T}$ by using
loss function in Eq. (6)(in paper). It can be shown that the difference
of the true action-value function and the estimated one through successor
feature functions and reward mapper, is bounded as:
\[
\mathrm{Pr}\Big(\Big|{Q_{\mathcal{T}}}^{\pi_{j}^{*}}(s,a)-\tilde{Q}_{\mathcal{T}}^{\pi_{j}^{*}}(s,a)\Big|\leq\varepsilon(m)\ \forall s,a\Big)\geq1-\delta,
\]
where \\
$\varepsilon(m)=\sqrt{2\mathrm{log}(2u_{m}/\delta)+8\mathrm{log}(2mbr\sqrt{\mathrm{log}(4a/\delta)})}$\ \\$\sigma_{m,d}([\boldsymbol{\mathrm{x}}]_{m})+\frac{1}{m^{2}},\ {\bf \boldsymbol{\mathrm{x}}\in\mathbb{X}}\ \delta\in(0,1)$,
$u_{m}=\frac{\pi^{2}m^{2}}{6}$, $m>1$ being the number of observations
in environment $\mathcal{T}$, $[\mathrm{\boldsymbol{\mathrm{x}}}]_{m}$
is the closest points in $\mathbb{X}_{m}$ to ${\bf \boldsymbol{\mathrm{x}}\in\mathbb{X}}$.
$\sigma_{m,d}(.)$ is the square root of posterior variance as defined
in Eq. (10)(in paper). $a,b>0$ are constants.

\textbf{Proof:} 
As explained in Section \ref{subsec:Infinite}, we know:
\small
\begin{multline}
\mathrm{Pr\Big(}\forall\mathrm{\boldsymbol{x}\in\mathbb{X}},|\psi_{\mathcal{T},d}^{\pi_{j}^{*}}(\mathrm{\boldsymbol{\mathrm{x}}})-\psi_{\mathcal{T},d}^{\pi_{j}^{*}}(\mathrm{\boldsymbol{\mathrm{x}}^{\prime}})|<b\sqrt{\mathrm{log}(4a/\delta)}\Big)|\mathrm{\boldsymbol{\mathrm{x}}-\mathrm{\boldsymbol{\mathrm{x}}^{\prime}}}|.\label{eq:Dis-1}
\end{multline}
\normalsize
Accordingly, by replacing $\mathrm{\boldsymbol{\mathrm{x}}^{\prime}}$:
\[
\mathrm{}\forall\mathrm{\boldsymbol{\mathrm{x}}\in\mathbb{X}}_{m},|\psi_{\mathcal{T},d}^{\pi_{j}^{*}}(\mathrm{\boldsymbol{\mathrm{x}}})-\psi_{\mathcal{T},d}^{\pi_{j}^{*}}([\mathrm{\boldsymbol{\mathrm{x}}}]_{m})|\leq2rb\sqrt{\mathrm{log}(4a/\delta)}/\tau_{m}.
\]

By selecting the discretisation factor as $\tau_{m}=4m^{2}br\sqrt{\mathrm{log}(4a/\delta)}$:
\[
\mathrm{}\forall\mathrm{\boldsymbol{\mathrm{x}}\in\mathbb{X}}_{m},|\psi_{\mathcal{T},d}^{\pi_{j}^{*}}(\boldsymbol{\mathrm{x}})-\psi_{\mathcal{T},d}^{\pi_{j}^{*}}([\boldsymbol{\mathrm{x}}]_{m})|\leq\frac{1}{m^{2}}.
\]
This implies $|\mathbb{X}_{m}|=(4m^{2}br\sqrt{\mathrm{log}(4a/\delta)})^{2}$.
By replacing $|\mathbb{X}_{m}|$ in $\beta_{m}$ defined in Section
\ref{subsec:Infinite}, the proof is completed. $\spadesuit$

Hence, if $\mathbb{X}$ is infinite set, Theorem $2$ holds with $\varepsilon(m)=\sqrt{2\mathrm{log}(2u_{m}/\delta)+8\mathrm{log}(2mbr\sqrt{\mathrm{log}(4a/\delta)})}\sigma_{m,d}([\boldsymbol{\mathrm{x}}]_{m})+\frac{1}{m^{2}}$.

\begin{figure*}[t]
\centering{}\includegraphics[scale=0.26]{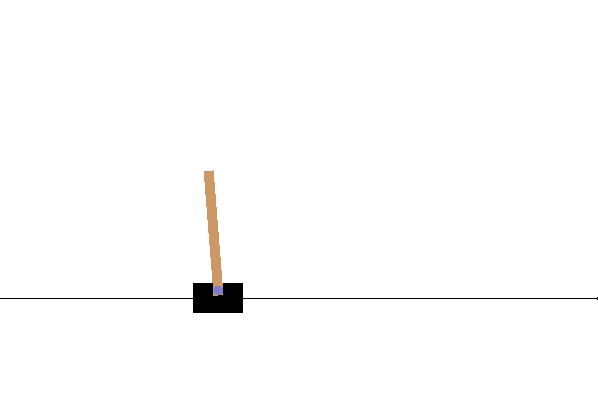} \includegraphics[scale=0.26]{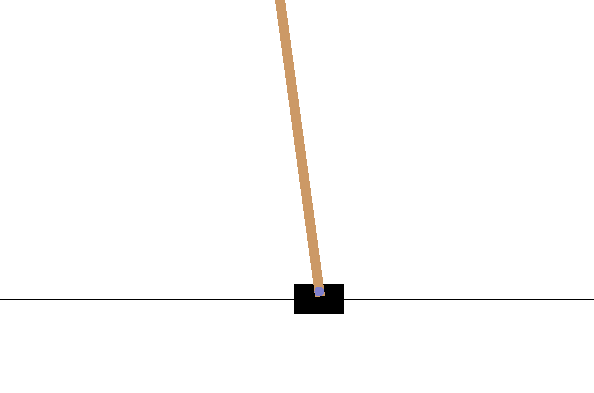}\caption{Illustration of the change in the dynamics for the CartPole problem.
(Left) Pole's length is $0.5m$ in the source environment and (Right)
Pole's length changed to $3m$ in the target environment.}
\label{fig:App_1}
\end{figure*}

\section{Experimental Details}

\paragraph*{Maze (navigation problem): }
For the task of navigation, we designed a maze environment with following
properties: (1) We set the $\varepsilon$-greedy exploration rate
to $\varepsilon_{e}=0.5$ for the adaptation phase with decay rate
of $0.9999$, this value is set to zero in the testing phase. (2)
Discount factor value is set to $\gamma=0.9$. (3) $\alpha=0.05$
is the learning rate. Agent is allowed to reach to the goal in maximum
of $100$ steps, otherwise it terminates.

\begin{figure}[t]
\begin{raggedright}
\begin{minipage}[t]{0.5\textwidth}%
\noindent \begin{center}
\includegraphics[width=0.7\textwidth]{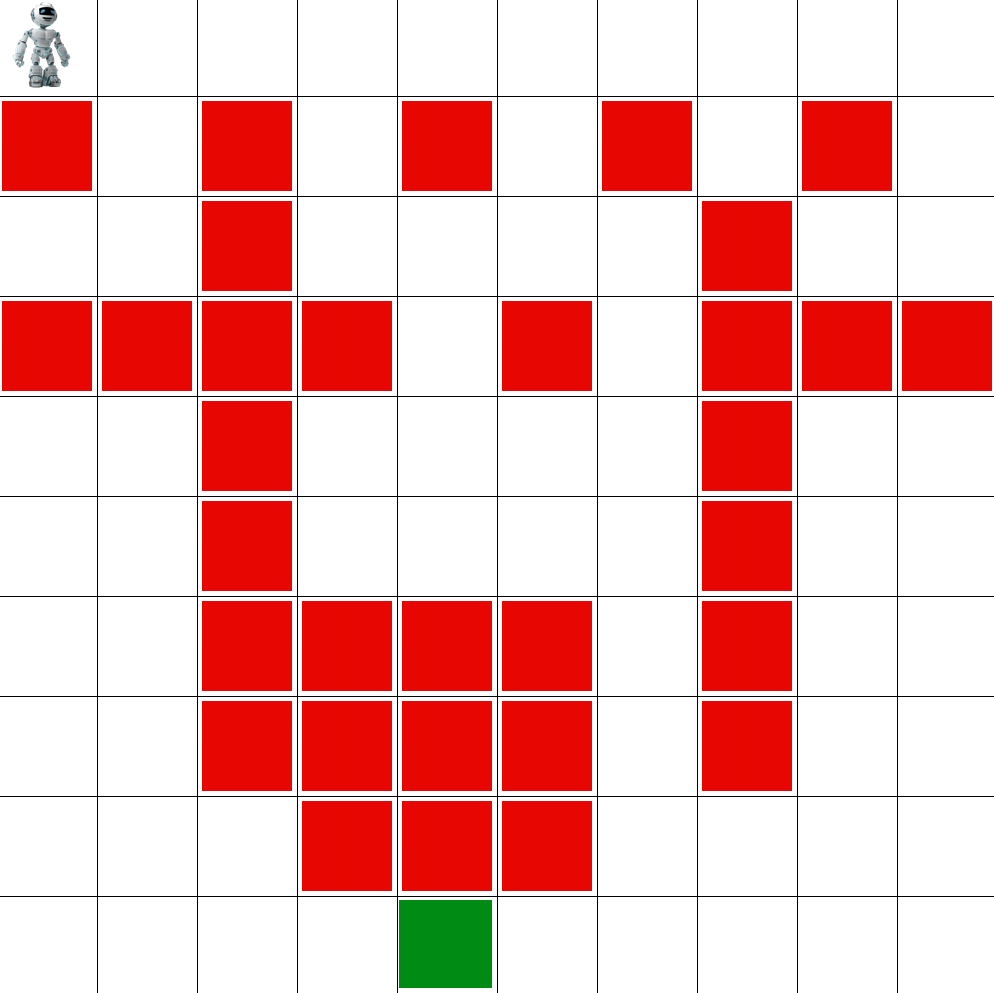}
\par\end{center}%
\end{minipage}
\par\end{raggedright}
\raggedright{}\caption{Illustration of the maze environment.\label{fig:Majid_Exp}}
\end{figure}

For $12$ source environments, $25$ obstacles are randomly generated,
the agent always start from top left, and the goal is also randomly
placed in these environments. We used generic Q-learning with replay
buffer size $10^{4}$ and Adam optimizer with batch size $64$ to
find the optimal policies in all these $12$ environments. Algorithm
1 in the paper is then used to estimate the successor feature functions
in the environment. Figure \ref{fig:Majid_Exp} demonstrates our toy
environment with agent at the top left, red obstacles, and the green
goal. Our proposed feature function for this problem is a MLP with
4 hidden layers with a linear activation function in the last hidden
layer to represent the reward mapper of the task. The remaining hidden
layers have ReLU activation function. The output of this network is
the predicted value of the reward for a state and action. Note that
this network minimises the loss function introduced in Eq. (6)(in
paper). We used SE kernel and maximising the log marginal likelihood for finding the best set of hyperparameters for GP.

\paragraph*{CartPole: }
As mentioned, to translate the image data into states, we used a CNN
with: (1) First hidden layer with $64$ filters of $5\times5$ with
stride $3$ with a ReLU activation, second hidden layer with $64$
filters of $4\times4$ with stride $2$ and a ReLU activation, third
hidden layer with $64$ filters of $3\times3$ with stride $1$ and
a ReLU activation. The final hidden layer is ``features'' we used
in an image that is fully connected Flatten units.

Maximum number of steps for the CartPole problem is set to $200$. We
set the $\varepsilon$-greedy exploration rate to $\varepsilon_{e}=0.5$
for the adaptation phase with decay rate of $0.9999$, this value
is set to zero in testing phase. The source learned policy is with
pole's length of $0.5m$ and accordingly, using Algorithm 1 in the
paper the corresponding successor features are extracted. We used
generic Q-learning with replay buffer size $10^{5}$ and Adam optimizer
with batch size $64$ to find this optimal policy. The target environment
is then modified to incorporate the change of environment. Figure
\ref{fig:App_1} demonstrates the change of dynamics. We used the
same structure of feature function in Maze problem for this experiment. 

\begin{figure}[t]
\begin{raggedright}
\begin{minipage}[t]{0.5\textwidth}%
\noindent \begin{center}
\includegraphics[width=0.6\textwidth]{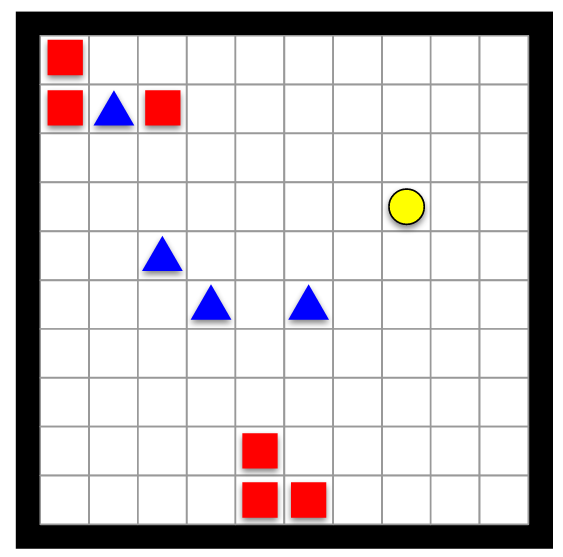}
\par\end{center}%
\end{minipage}
\par\end{raggedright}
\raggedright{}\caption{An example of the environment described in FSF \cite{barreto2020fast}.\label{fig:An-example-of}}
\end{figure}

\paragraph*{FSF:}
This environment \cite{barreto2020fast} is a $10\times10$ grid
with $10$ objects and an agent occupying one cell at each time step.
There are $2$ types of objects each with a reward associated with
it. We randomly initialise those $10$ objects by sampling both their
type and position from a uniform distributions over the corresponding
sets. Likewise, the initial position of the agent is a uniform sample
of all possible positions in the grid. The reward function is defined
by the object type that has been picked up by the agent. e.g. Picking
up red object is $+1$ reward and picking up blue is $-1$. Agent
picks up an object if it occupies that particular cell in which the
object exists. If agent picks up an object, another one will be generated
randomly (in terms of location and type) in the grid. At each step
the agent receives an observation representing the configuration of
the environment \cite{barreto2020fast}. These are $11\times11\times(\mathbb{D}+1)$
tensors that can be seen as $11\times11$ images with $(\mathbb{D}+1)$
channels that are used to identify objects and walls \cite{barreto2020fast}.
The observations are shifted so that the the agent is always at the
top-left cell of the grid. Figure \ref{fig:An-example-of} shows an
example of this environment. The two source policies used in this
experiment have $\mathrm{\boldsymbol{w}_{1}=[1,0]},\mathrm{\boldsymbol{w}_{2}=[1,0]},$
respectively. Intuitively, the reward mappers indicate picking up
an object of particular type and ignoring the other type. However,
in the target environment, the change of reward function is to pick
up the first object type and ``avoid'' the second one with negative
reward - i.e. $\boldsymbol{w}_{\mathcal{T}}=[1,-1].$ For the dissimilarity
of dynamics, we added $5\%$ noise to the transitions of the agent
and also randomly placed a terminal state in the target environment
with -1 reward.

\subsection{Additional Experiments}

\begin{figure}[t]
\begin{center}
\includegraphics[scale=0.45]{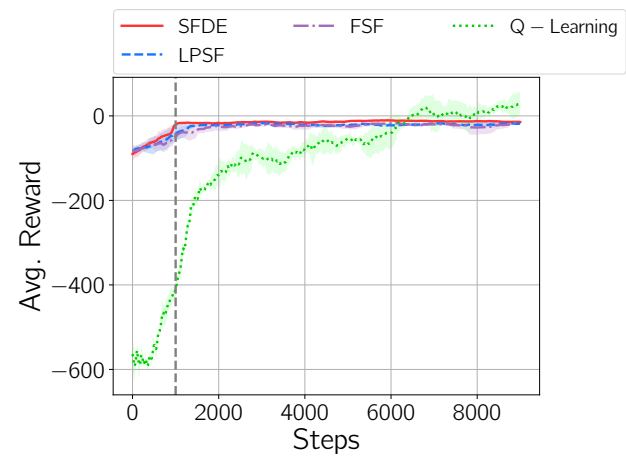}
\end{center}
\caption{Performance of Q-Learning in navigation problem. \label{fig:QL}}
\end{figure}

In this section, we compare our results in navigation problem with
generic Q-Learning. We relaxed the assumption adaptation and testing
phase and Q-Learning is allowed to use all the observations from the
target environment. Figure \ref{fig:QL} shows Q-Learning also converges
to the same amount of avg. reward, however, since no transfer is involved, it is significantly slower than other baselines. For Q-Learning,
we set $\varepsilon_{e}=0.9$ with a decay rate of $0.9999$ in $10^{4}$
iterations.

\end{document}